\documentclass[twocolumn]{bmcart}
\usepackage{amsthm,amsmath}
\usepackage{booktabs}
\usepackage{multirow}
\usepackage{graphicx}
\usepackage{url}
\usepackage{longtable}

\usepackage{soul}
\usepackage{linguex}
\usepackage[utf8]{inputenc} 

\startlocaldefs
\endlocaldefs

\begin{document}

\begin{frontmatter}

\begin{fmbox}
\fcolorbox{black}{yellow!40}{\begin{minipage}{47.6em}
This is the extended journal paper version of our IEE ICHI 2020 paper `Information Extraction Models for German Clinical Text'. Since then this version is still under review ...
\end{minipage}
}
\medskip

\dochead{Research}


\title{A Medical Information Extraction Workbench to Process German Clinical Text}


\author[
  addressref={aff1,aff2},                   
  corref={aff1},                       
  email={roland.roller@dfki.de}   
]{\inits{R.R.}\fnm{Roland} \snm{Roller}}
\author[
  addressref={aff1},
  email={laura.seiffe@dfki.de}
]{\inits{L.S.}\fnm{Laura} \snm{Seiffe}}
\author[
  addressref={aff1},
  email={ammer.ayach@dfki.de}
]{\inits{A.A.}\fnm{Ammer} \snm{Ayach}}
\author[
  addressref={aff1},
  email={sebastian.moeller@dfki.de}
]{\inits{S.M.}\fnm{Sebastian} \snm{M{\"o}ller}}
\author[
  addressref={aff1},
  email={oliver.marten@dfki.de}
]{\inits{O.M.}\fnm{Oliver} \snm{Marten}}
\author[
  addressref={aff1,aff2},
  email={michael.mikhailov@charite.de}
]{\inits{M.M.}\fnm{Michael} \snm{Mikhailov}}
\author[
  addressref={aff1},
  email={christoph.alt@dfki.de}
]{\inits{C.A.}\fnm{Christoph} \snm{Alt}}
\author[
  addressref={aff3},
  email={danilo.schmidt@charite.de}
]{\inits{D.S.}\fnm{Danilo} \snm{Schmidt}}
\author[
  addressref={aff2},
  email={fabian.halleck@charite.de}
]{\inits{F.H.}\fnm{Fabian} \snm{Halleck}}
\author[
  addressref={aff2,aff4},
  email={marcel.naik@charite.de}
]{\inits{M.N.}\fnm{Marcel} \snm{Naik}}
\author[
  addressref={aff2,aff4},
  email={wiebke.duettmann@charite.de}
]{\inits{W.D.}\fnm{Wiebke} \snm{Duettmann}}
\author[
  addressref={aff2},
  email={klemens.budde@charite.de}
]{\inits{K.B.}\fnm{Klemens} \snm{Budde}}



\address[id=aff1]{
  \orgdiv{Speech and Language Technology},             
  \orgname{German Research Center for Artificial Intelligence (DFKI)},          
  \city{Berlin},                              
  \cny{Germany}                                    
}
\address[id=aff2]{%
  \orgdiv{Department of Nephrology and Medical Intensive Care},
  \orgname{Charité – Universitätsmedizin Berlin},
  \city{Berlin},
  \cny{Germany}
}
\address[id=aff3]{%
  \orgdiv{Business Unit IT},
  \orgname{Charité – Universitätsmedizin Berlin},
  \city{Berlin},
  \cny{Germany}
}
\address[id=aff4]{%
  \orgdiv{Participant in the digital clinician scientist programme},
  \orgname{Berlin Institute of Health},
  \city{Berlin},
  \cny{Germany}
}





\begin{abstractbox}

\begin{abstract} 
\parttitle{Background} 
In the information extraction and natural language processing domain, accessible datasets are crucial to reproduce and compare results. Publicly available implementations and tools can serve as benchmark and facilitate the development of more complex applications. However, in the context of clinical text processing the number of accessible datasets is scarce -- and so is the number of existing tools. One of the main reasons is the sensitivity of the data. This problem is even more evident for non-English languages. 
\parttitle{Approach} 
In order to address this situation, we introduce a workbench: a collection of German clinical text processing models. The models are trained on a de-identified corpus of German nephrology reports.

\parttitle{Result} 
The presented models provide promising results on in-domain data. Moreover, we show that our models can be also successfully applied to other biomedical text in German. Our workbench is made publicly available so it can be used out of the box, as a benchmark or transferred to related problems.

\end{abstract}




\end{abstractbox}
\end{fmbox}

\end{frontmatter}

\section*{Background}

The year is 2022 AD. The research community relies entirely on neural and computationally intensive methods using large datasets and pre-trained language models, which push the current development rapidly forward. Well, not entirely. One small domain still struggles with more fundamental problems: lack of existing datasets, corpora or pre-trained models; hurdles with legal aspects; and outdated computer infrastructure with limited user rights -- just to name a few. The domain is called clinical text processing. This applies particularly for non-English clinical text processing. 


In the English speaking world, however, various tools can be applied to process clinical text, such as cTAKES \cite{savova2010mayo} or MetaMap \cite{aronson2010overview}. Most of those tools focus on named entity recognition, partly combined with concept normalization and disambiguation. Moreover, some tools target a very particular topic or domain, such as pharmacovigilance (MedLEE \cite{friedman1994general}),  detection of medications (MedEx \cite{xu2010medex}) or temporal information (MedTime \cite{sohn2013comprehensive}). A more detailed overview can be found in Wang et al. \cite{WANG201834}. 

Regarding the availability of annotated clinical text in English, various challenges have been conducted in the last years. Challenges that share data and experiences on the same problem. The i2b2 NLP challenge (now n2c2) for instance addresses a large variety of different tasks, such as medication detection \cite{uzuner2010extracting} or identification of heart disease risk factors \cite{stubbs2015identifying}. Also other shared tasks such as CLEF eHealth or SemEval have been carried out already various times on clinical text (e.g. \cite{suominen2013overview,kelly2014overview,pradhan2014semeval,bethard2016semeval}). In more recent years, those challenges also targeted non-English clinical texts. The CLEF eHealth challenge 2018 (Task 1 \cite{neveol2018clef}), for instance, focused on the mapping of ICD codes to death certificates in French, Hungarian, and Italian, or NEGES at IberLEF targets negation detection from Spanish clinical reports \cite{jimenez2019neges}. A good overview about non-English clinical text processing in general is provided in N{\'e}v{\'e}ol et al. \cite{neveol2018clinical}.

For the German language, the situation looks worse in comparison to French or Spanish. In 2019 a CLEF eHealth challenge for German text had been carried out, in which ICD-10 codes were mapped to health-related, non-technical summaries of experiments \cite{dorendahl2019overview}. However, this data concerned animals, not humans. Besides, not much other data has been published. An overview of the current situation is presented in Borchert et al. \cite{borchert2020ggponc}. The paper lists 13 different German text corpora with a clinical/biomedical context, but only three are freely available. First, GGPOnc \cite{borchert2020ggponc} a dataset of clinical practice guidelines, second TLC \cite{seiffe2020witch}, posts of a patient forum with annotated laymen expressions, and finally JSynCC \cite{lohr2018sharing} a dataset of German case reports extracted from medical literature. In the case of JSynCC, authors actually provide a software to extract the relevant text passages from digital medical books, instead of providing the data itself -- due to legal reasons.

Concerning existing tools and pre-trained methods to process German clinical text, the situation is similar to the availability of text data. Most prominent is JPOS \cite{Hellrich2015}, a tokenizer and part of speech tagger trained on clinical text. The authors published the tool, as they were not allowed to publish the underlying FRAMED corpus \cite{wermter2004really} itself. In addition to that, two NegEx \cite{CHAPMAN2001301} versions for German exist \cite{chapman2013extending} \cite{cotik-etal-2016-negation}, as well as a dependency tree parser \cite{Kara:2018}, an abbreviation expansion \cite{oleynik2017unsupervised} and a tool to pseudonomize protected health information (PHI) in German clinical text  \cite{lohr2021pseudonymization}. 


In very recent times, some additional resources highly related to this work have been published, as the field is developing quickly in recent years. BRONCO \cite{kittner2021annotation}, is an annotated dataset of German discharge summaries of the oncology domain - the first clinical text dataset in German which has been published. The data is de-identified and sentences of all 200 documents were shuffled to lower the risk of any re-identification. Along with the data, the authors also make the baseline models available on request. GERNERMED \cite{frei2022gernermed} is a German medical NER model, which is trained on translated data from n2c2 2018 \cite{henry20202018}. And finally, German MedBERT \cite{shrestha2021development}, a BERT model, optimized for German clinical text, has been published on Hugging Face\cite{HFGermanMedBERT}, targeting ICD10 code mapping. 

In order to further support the development of resources to process German clinical text, this work describes an annotated dataset of German nephrology reports. It contains fine-grained annotations of concepts, relations, attributes, as well as part-of-speech (POS) labels and dependency trees. Unfortunately, we are unable to share the dataset at this point due to unsolved data protection concerns. Thus, instead of publishing the dataset, we release the machine learning models which we trained on the de-identified data\footnote{The workbench can be found here: \url{http://biomedical.dfki.de} and \url{https://github.com/DFKI-NLP/mEx-Docker-Deployment}. In order to use the models, a user agreement has to be signed first.}. While the BRONCO models mainly target diagnosis, treatment, medications, as well as factuality, our work includes a larger variety of different named entities, which might be useful for other use cases. Moreover, our workbench also includes relation detection and a POS tagger. Similarly this applies to GERNERMED, which mainly targets medications, as well as its dosage, duration, frequency etc. Note, as this work was developed for over various years, we still rely on classical word embeddings, rather than testing the efficiency of German MedBERT for our scenario.

\begin{table*}[ht!]
  \centering
  \small
  \caption{Overview of all concepts}
\begin{tabular}{llp{6.2cm}p{5cm}}
\toprule
    & \textbf{Concept} & \multicolumn{1}{c}{\textbf{Description}} & \multicolumn{1}{c}{\textbf{Translated examples}} \\
    \midrule
    \multirow{5}{*}{\rotatebox[origin=c]{90}{\textbf{Central}}} & Medical Condition & Signs, symptoms, diagnoses, diseases, findings & \textit{Aphasia, cachectic} \\
    & Diagnostic Lab Procedure & Procedures (diagnostic or laboratory) that serve the clinical examination of the patient's state &  \textit{measure, CT angiography, sonography}\\
    & Treatment & All variants of clinical interventions aiming at improving the health state & \textit{blood pressure management, transplantation} \\
    \midrule
    \multirow{12}{*}{\rotatebox[origin=c]{90}{\textbf{Relating}}} & Medication & Names of medications, their active substances & \textit{Prograf, Sandimmun} \\
    & Biological Chemistry & Biochemical substances that play a role in human organism & \textit{creatinine, ANA, HbA1c} \\
    & Process & Endogenous processes and functions in human organsim & \textit{peristaltic sounds, defecation} \\
    & Person & Mentions of people & \textit{gynecologist, GP} \\
    & Body Part & Parts of the human body & \textit{renal, lung} \\
    & Body Fluid & Fluid and excretions of the human body & \textit{urine, sputum, blood} \\
    & Medical Device & Artificial or biological system that supports or replaces a failed function of human organism & \textit{kidney allograft, TEP, shunt} \\
    & Biological Parameter & Functions, features and characteristics of the human body & \textit{nutritional status, blood count, body weight} \\
    \midrule
    \multirow{8}{*}{\rotatebox[origin=c]{90}{\textbf{Specifying}}} & Medical Specification & Clinically specifying elements & \textit{chronic, symptomatic} \\
    & Local Specification & Locally specifying elements & \textit{perirenal, right} \\
    & Time Information & Temporally specifying elements & \textit{today in the morning, 2014, on 11.01.2018}\\
    & Dosing & Dosing instructions for medications & \textit{1 sachet daily, 1 in the morning - 1 in the afternoon - none at night, 5 mg daily} \\
    & Measurement & Measurements and evaluations of functions, findings, states & \textit{sonorous, active, distinct} \\
    & State of Health & The (aimed) health state or the ongoing improvement & \textit{properly adjusted, satisfactory, stable} \\
    \bottomrule
    \end{tabular}
  
\label{conceptschema}
\end{table*}

\begin{table*}[ht!]
  \centering
  \small
  \caption{Overview of all relations}
\begin{tabular}{ llp{13cm} }
\toprule
\textbf{} & \textbf{Relation} & \textbf{Description} \\
\midrule
\multirow{8}{*}{\rotatebox[origin=c]{90}{\textbf{Describing}}} & Has\_state & \textit{Argument1} is described as being pathologic (\textsc{Medical\_Condition}) or as being healthy (\textsc{State\_of\_health}) \\ 
& Has\_dosing & A \textsc{Medication} or a \textsc{Treatment} is linked to a \textsc{Dosing} instruction \\
& Has\_time\_info & \textit{Argument1} is described by a \textsc{Time\_information} \\ 
& Has\_measure & \textit{Argument1} is described by a \textsc{Measurement} \\
& Is\_located & \textit{Argument1} is described locally by \textsc{Local\_specification} or \textsc{Body\_part} \\
& Is\_specified & \textit{Argument1} is described by \textsc{Medical\_specification} \\
\midrule

\multirow{3}{*}{\rotatebox[origin=c]{90}{\textbf{Medical}}} & Shows & A \textsc{DiagLab\_Procedure} or a \textsc{Biological\_Parameter} leads to a finding (\textit{Argument2}) \\
& Examines & A \textsc{DiagLab\_Procedure} examines \textit{Argument2}  \\
& Involves & A \textsc{Treatment} involves a \textsc{Medical\_Device}, a \textsc{Medication} or another \textsc{Treatment}  \\

\bottomrule
\end{tabular}
\label{annotated_relations}
\end{table*}

\section*{German Nephrology Corpus}\label{corpus}%

The corpus consists of German documents of the nephrology division at Charité – Universitätsmedizin Berlin. All documents have been de-identified by removing protected health information (PHI) defined by HIPPA (Health Insurance Portability and Accountability Act), using deID \cite{Seuss2017}. Next, the documents were enriched with semantic annotations as described in the following. 

We considered two different document types for our annotations,\ \textit{clinical notes} and\ \textit{discharge summaries} (or \textit{discharge letters}). Both document types report on kidney transplanted patients who underwent long-term treatment, are written by medical professionals and address medical professionals.\ \textbf{Discharge summaries} (``Arztbriefe") serve as a summary of a patient's hospital stay, expressed as letters sent to the patient's GP, and cover history, diagnostic and therapeutic procedures. Since discharge letters are relatively long, they often provide additional structural elements such as headings, enumerations, etc.\ \textbf{Clinical notes} (``Verlaufsnotizen") summarize the results from a single consultation in the outpatient department, which results in rather short texts. In contrast to discharge letters, their content, form, and function are not subject to professional and structural standards, as they are used only for internal communication and largely depend on the doctor's writing style.

\begin{table*}[ht!]
  \centering
  \small
  \caption{Overview of all attributes}
\begin{tabular}{ llp{12cm} }
\toprule
\textbf{Attribute} & \textbf{Value} & \textbf{Explanation} \\
\midrule
DocTime & Past & The entity or the event existed or happened in the past, respectively (related to the current temporal setting of the document) \\
& Future & The entity or the event is planned, prescribed or recommended in the future (related to the current temporal setting of the document) \\
& Past\_present & The entity or the event has begun in the past and endures to the current temporal setting of the document \\


\midrule
LevelOfTruth & Possible\_future & The entity or the event might happen in the future, e.g. in an if-clause \\
& Negative & The entity or the event is negated \\
& Speculated & The truth of the entity or the event is only assumed\\
& Unlikely & The truth of the entity or the event is doubtful \\
\bottomrule
\end{tabular}

\label{annotated_attributes}
\end{table*}

\subsection*{Characteristics of Clinical Language}
A characteristic of the (German) clinical language is the large proportion of technical terms, which mostly have their origin in Latin or Greek, and underlie specific morphological rules. Moreover, clinical language provides a characteristic and individual use of syntax, for example, a notation style that excludes function words (e.g., articles, auxiliary verbs) and might include non-standard term variants. As documents might be written in a hurry, sentences can include typos, can be incomplete or punctuation marks are missing. Overall, the clinical language prefers a compact and reduced language use with a high information density. Example~\ref{notation} shows a typical sentence structure of a clinical note.


\ex. Im Sono kein Stau.\\\textit{Sonogram [shows] no congestion.}\label{notation}

This short expression does not use a verb - something like ``to show'' is presumably meant in this context. This information is however significant as it forms the relation between the examination process (\textit{Sono},  ``Sonogram") and the (negated) finding (\textit{Stau}, ``congestion"). As the verb and therefore the relation is only implicitly expressed, a basic understanding of the subject is presupposed for the correct manual annotation. 

Furthermore, many abbreviations are used in both document types. These abbreviations are often standardized, and their expansion and meaning are well documented. Conversely, the expansion of abbreviations can be complicated by the fact that abbreviations are often ambiguous. For example the German web dictionary \textit{Beckers Abkürzungslexikon Medizinischer Begriffe}\cite{Beckers} for medical-related abbreviations lists 61 possible expansions for \textit{KS} (for instance \textit{Kaltschweißigkeit, Klopfschall, Kaiserschnitt, Kaufmann-Schema}) with additional subcategories of varieties. Only a sufficient context can help to disambiguate an abbreviation. However, as the clinical language tends to a compact and reduced language, such a disambiguating context is not always given. Another characteristic of clinical documents is the large number of negations and vague descriptions, particularly in context of symptoms and findings.

\subsubsection*{Semantic Annotations}
Our semantic annotation schema is intended to cover the most relevant textual information in the corpus, and developed during the annotation process \cite{Roller:2016}. 
The schema has been developed from scratch, together with linguists, computer scientists, and physicians and focuses on the pathological health state (medical condition) of the patient as well as his or her treatments and diagnostic and laboratory examinations. The schema targets mainly the recognition of that information and everything which is connected to it. It is individually adapted to the demands of the German nephrology domain and applies for both discharge summaries and clinical notes. In order to gather this meaning, the schema is constructed of concepts, binary relations, and concept attributes which are introduced in the following.

\textbf{Concepts}: The concept schema can be divided into three groups: \textbf{central}, \textbf{relating}, and \textbf{specifying}. \textbf{Central} concepts describe -- from our perspective -- the most crucial information about a patient. It concerns the pathological health state of the patient as well as his or her treatments and diagnostic and laboratory examinations. \textbf{Relating} concepts describe other relevant information within the documents. By connecting them via relations (see below) to mostly central concepts, those information help to gather relevant information of the documents. \textbf{Specifying} concepts provide more detailed information to the other concepts, such as dosing, local or time information. An overview of the concept schema is provided in Table \ref{conceptschema} and includes a short definition and examples. 

\textbf{Relations}: Our relation schema describes a binary semantic relation between two concepts. It intends to connect the annotated concepts within the document with each other and to give the single concepts a stronger meaning. On a high level, relations can be divided into two groups, \textbf{describing} and \textbf{medical} relations. \textbf{Describing} relations connect two concepts with each other, of which one argument adds more information to the other one, such as the dosing of a medication, the pathological state of something, or a further specification. Usually one argument is a \textbf{specifying} concept. \textbf{Medical} relations instead describe more complex situations related to the examination and treatment of a patient. 

Table \ref{annotated_relations} presents the set of relations including a short description. In most cases the relations are defined in a broader sense. While one argument is usually defined to be bound to one or two particular concept types, the other argument often has more freedom and can be bound to various concept types.

\textbf{Attributes}: Attributes are used for the further specification of an annotated concept. While a concept covers the term's lexical information, the selected attribute value refers to extra-lexical/contextual information. Such information relates to \textbf{temporal information} 
or information about the \textbf{level of truth}. See Table \ref{annotated_attributes} for the annotation schema of attributes.

The time information attribute \textit{DocTime} helps to structure the described temporal course of the document. By applying one of its values, an entity can be highlighted as has happened in the past or as being planned in or predicted for the future. In most cases, a surrounding \textsc{time\_information}-concept triggers the attribute selection. If possible, both concepts are additionally linked by the relation \textsc{has\_time\_info}. This means that in many cases, this information is expressed twice: First by concept and second by attribute value. 

The attribute \textit{LevelOfTruth} highlights information that indicates vagueness, possibility, and negated expressions. Generally, both document types comprise plenty of expressions of assumptions. It is necessary to differentiate between a statement expressing certainty and an assumption, for example.

\subsection*{Corpus Generation}\label{annotation_proc}

The annotation was carried out 
by three students (2 linguists, 1 medical student). The medical student in particular contributed to the understanding of the medical terminology. The task itself was conducted by using the Brat annotator tool \cite{Stenetorp:2012} within several annotation cycles. This method led to various adaptations and updates of the annotation schema.

\begin{table}[h]
  \centering
  \caption{Analysis of annotated documents}
\begin{tabular}{ lcc }
\toprule
 & \textbf{Discharge Summaries}  & \textbf{Clinical Notes} \\
\midrule
\textbf{\# docs} & 61 & 1300  \\
\textbf{\# words} & 57,219 &  54,206  \\
\textbf{\# sentences} & 6,213 & 6,618   \\
\textbf{avg. words (std)} & 938 (246.33) &  54 (45.43)  \\
\bottomrule
\end{tabular}

\label{first_analysis_corpus}
\end{table}

Table \ref{first_analysis_corpus} provides an overview of the annotated dataset. Overall 1300 clinical notes and 61 discharge summaries have been annotated. Most documents were examined at least twice by two different annotators.
However, between the two linguists, which annotated more than 80\% of the data (53.3\%, 30.9\%), there exists a larger number of overlapping documents in comparison to the medical student (11\% between linguists, 8\% and 2\% between the linguists and the medical student; only a small portion of documents was annotated by all three annotators, namely 1 discharge summary and 20 clinical notes). 

The table shows the number of different annotated documents, the overall number of words in all documents, the overall number of sentences and the average
words per document, including standard deviation (in brackets).
Discharge summaries contain a larger average number of words per document compared to the clinical notes\footnote{The information is generated by applying a German tokenizer and a sentence splitter.}. However, the standard deviation of the average word number per document shows that both document types have a large variation in text length. Some clinical notes contain only a few words.

Similarly as in Hripcsak and Rothschild \cite{hripcsak2005agreement} we calculate the inter-annotator-agreement (IAA) using the pairwise average  F-score (micro) on character level. This results in an avg. F-Score across all annotators of 0.761 for the concepts and 0.636 for relations. 
Particularly the score of the relations does not seem to be very high. However, two aspects need to be taken into consideration: a consistent relation annotation strongly depends on the fact if the underlying concepts are annotated correctly beforehand, and secondly, the IAA between the two linguists, who annotated more than 80\% of the data, have got a much stronger overlap, namely 0.822 for concepts and 0.697 for relations. 



\subsection*{Challenges \& Limitations}\label{challenges}

In order to meet the complexity of the German clinical language, we created a detailed and extensive annotation schema. We faced multiple challenges during the annotation process, due to the complexity of the language and the schema. These challenges had a substantial impact on the consistency of the annotations across the annotations (IAA), therefore decisions about the approach had to be made. 
The most relevant ones will be presented in the following.

\textbf{Fine-Grained Annotation:} The German language includes a large number of compound words. As compounds consist of two (or more) meaningful units that are linked via an inherent linguistic relation, the annotation of that relation seems possible. This can lead to subword annotations on multiple levels. Notably this applies to medical technical terms.

\begin{figure}[bht!] 
  \centering
     \scalebox{1.5}{\includegraphics{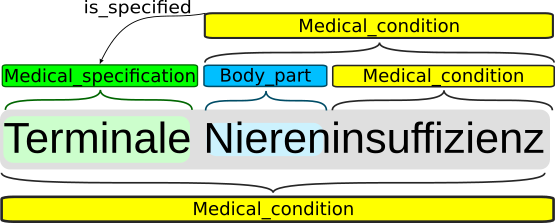}}
   \caption{Annotation Granularity}
  \label{fig:annotation_granularity}
\end{figure}

\begin{figure}[bht!] 
  \centering
		\scalebox{1.2}{\includegraphics{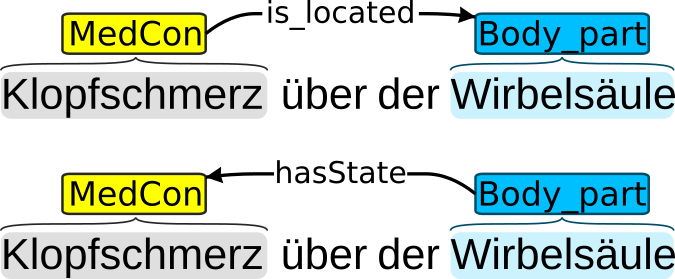}}
		\caption{``Pain on percussion along the spine"}
          \label{fig:direct1}
\end{figure}

Figure \ref{fig:annotation_granularity} shows an example of annotating on different levels. It shows the compound \textit{Niereninsuffizienz} (``renal insufficiency") whose two elements refer to the concepts \textsc{body\_part} and \textsc{medical\_condition}. The word itself also refers to a \textsc{medical\_condition}. The specifying adjective \textit{terminale} (``terminal") should be considered as an attached part of the technical term. Thus the preferred annotation here is \textit{Terminale Niereninsuffizienz} as \textsc{medical\_condition}. Alternatively, \textit{terminale} is annotated as \textsc{medical\_specification}, \textit{Nie- reninsuffizienz} as  \textsc{medical\_condition} and both are connected via the relation \textsc{is\_specified}. The process of subword annotation goes beyond the scope of this corpus, therefore we opt for the annotation of larger spans. However, the possibility of both alternatives decreases the consistency in the annotation.

\textbf{Ambiguity:} The relations in our annotation schema are rather broadly defined. This means that one relation can be used to link several different argument pairs. For example, the relation \textsc{shows} with the concept \textsc{medical\_condition} as the second argument can make use of two different first arguments: Firstly it links to \textsc{diagLab\_procedure}. In such a case, the result or the finding of a diagnostic procedure is expressed. Alternatively it links to \textsc{biological\_parameter}. Then the relation expresses that a specific parameter indicates a pathological condition. 

In some contexts, the connection between the two entities can be expressed equally by two different relations. This is mainly the case for the link between the two concepts \textsc{medical\_condition} and \textsc{body\_part}. See Figure \ref{fig:direct1}: The sentence describes the location of a symptom. This finding can either be described by using a \textsc{is\_located}-relation (Figure \ref{fig:direct1}, first line) or by defining the health state of a body part as being pathological (\textsc{has\_state}-relation, Figure \ref{fig:direct1}, second line). Both are, according to our schema, correct and express the intended semantic relation equally. In order to achieve a consistent annotation, we opt for the first version as the \textsc{medical\_condition}-information seems more central here.


\section*{Additional Datasets}\label{corpus}%

While the previous section introduced the main corpus in detail, this part presents additional relevant data sources, namely \textit{Nephro\_Gold}, the \textit{Hamburg Dependencies Dataset}, as well as a biomedical text corpus in German.


\subsection*{Nephro\_Gold: A syntactic dataset}

In addition to the semantic clinical corpus, also a small clinical syntactic dataset has been created in previous work \cite{Seiffe:2018} \cite{Kara:2018}. We refer to this corpus as \textit{Nephro\_Gold}. The dataset also consists of clinical notes and discharge summaries, is rather small (44 clinical notes and 11 discharge summaries) and is a subset of the dataset described above. It includes part-of-speech (POS) annotations using the Stuttgart-T\"ubingen Tagset (STTS) \cite{schiller1999guidelines} and dependency trees using the Universal Dependencies (UD) tagset \cite{nivre2016}. 


\subsection*{Hamburg Dependency Treebank}

The Hamburg Dependency Treebank (HDT) \cite{Foth:2014} is a large dataset of more than 261,000 German sentences, and includes POS labels and syntactical annotations. The annotation scheme of HDT is also based on the Stuttgart-T{\"u}bingen Tag Set for morphological and POS annotation and a set of 35 dependency labels for dependency annotation. The corpus is freely available for scientific purposes. In this work, the dataset will be used for the POS tagger.


\subsection*{BTC - A Biomedical Text Collection in German}

Many modern NLP (natural language processing) methods use word embeddings as input. However, it turns out that embeddings specialized on the given domain often outperform embeddings trained only on general text. Although embeddings with a focus on the biomedical domain have been published, there are no embeddings specialized on German clinical text\footnote{Note, after submission of the manuscript, the situation has slightly changed. See related work for more information.}. Moreover, as mentioned above, clinical text contains a large number of technical terms, which do not frequently occur in general news text or Wikipedia. This might result in many out-of-vocabulary words if we train a system using standard pre-trained embeddings. 

For this reason, we collected German text data from multiple sources in order to train our own custom embeddings. Table \ref{biomedical_text_collection} and Table \ref{medical_books} provide an overview about the different sources used. We scraped data from multiple webpages with a focus on biomedical topics, as well as forums. In addition to that we also used text from different medical books. In the following we will refer to our biomedical text collection as \textit{BTC}.

\begin{table*}[h!]
  \centering
\caption{Overview of different biomedical text sources in German to create a new biomedical text collection}
\begin{tabular}{ lll }
\toprule
\textbf{Size} & \textbf{Source} & \textbf{Description} \\
\midrule
7,5 GB & Med1 Forum\cite{Med1}  & German forum for clinical topics and information exchange   \\

91,8 MB &  Deutsches Medizin Forum\cite{DeutschesMedizinForum} &  German forum for clinical topics and information exchange  \\

28,6 MB & Spiegel Online\cite{Spiegel} & German news webpage, articles downloaded from the health section  \\


10,7 MB & Aerzte-Blatt\cite{AerzteBlatt}  & Official news publication of the German Medical Association  \\

10 MB & NetDoktor\cite{NetDoktor} & German online health portal for medical information from experts to patients   \\

7,1 MB & Onmeda\cite{Onmeda} & German online health portal, content extracted from the symptoms and diseases \\

3,6 MB & German PubMed Abstracts\cite{GermanPubMed}  & Archive of biomedical and life sciences journal literature  \\

1,9 MB & eDocTrainer\cite{eDocTrainer}  &  German collection of clinical case studies from all specialist disciplines  \\

16,5 MB & Medical Books &  Content from various medical books, see Table \ref{medical_books} \\

\bottomrule
\end{tabular}
\label{biomedical_text_collection}
\end{table*}

\begin{table*}[h]
  \centering
\caption{Overview of medical books}
\begin{tabular}{ lll }
\toprule
\textbf{Name} \textbf{Author}\\
\midrule
 \textit{Chirurgie: Mit integriertem Fallquiz - 40 Fälle nach neuer AO}. Springer-Verlag, 2009 \cite{siewert2009chirurgie} & Siewert, Stein\\
\textit{Neurologie}. Springer-Verlag, 2006 \cite{poeck2006neurologie} & Poeck, Hacke\\
\textit{Urologie}. Springer-Verlag, 2014 \cite{hautmann2014urologie} & Hautmann, Gschwend\\
\textit{Basiswissen Augenheilkunde}. Springer-Verlag, 2016 \cite{walter2016basiswissen} & Walter, Plange \\
\textit{Hals-Nasen-Ohren-Heilkunde}. Springer-Verlag, 2012 \cite{lenarz2012hals} & Lenarz, Boenninghaus \\
\textit{Notfallmedizin}. Springer-Verlag, 2016 \cite{ziegenfuss2016notfallmedizin} & Ziegenfuß \\
\textit{Basiswissen Dermatologie}. Springer-Verlag, 2017 \cite{goebeler2017basiswissen} &  Goebeler, Hamm\\ 
\textit{Basiswissen Psychiatrie und Psychotherapie}. Springer-Verlag, 2011 \cite{arolt2011basiswissen} & Arolt, Reimer, Dilling \\
\bottomrule
\end{tabular}

\label{medical_books}
\end{table*}

\section*{Methods and Setup}\label{methods_results}

This section provides an overview of the technical aspects of our work. We briefly present the relevant methods and explain how we use or modify them for our experiments. As we mainly rely on existing implementations, the technical components will be presented relatively short. 

\subsection*{Custom Word Embeddings with fastText}

All methods used in this work rely on variant types of word, character, and document embeddings. For reasons of simplicity and compatibility with our NLP pipeline, fastText \cite{bojanowski2016enriching} was the method of choice. fastText is a lightweight library that offers pre-trained text classifiers. Moreover, it provides various multilingual word vectors that can be fine-tuned on new unlabeled data to obtain a better domain-specific characterization of our clinical data. In this work we used fastText to fine-tune our own embeddings using BTC and the German Nephrology corpus. We refer to this representation as `custom embeddings'. A total of 5 epochs were needed using CBOW for the generation of the new embeddings. In addition, we used the default German fastText embeddings, which we refer to as `default'.



\subsection*{Clinical Text processing with Flair}
Large parts of our work relied on Flair \cite{akbik-etal-2018-contextual}, a state of the art NLP framework, which provides various functionalities, for instance, named entity recognition (NER) and part-of-speech-tagging (POS). Furthermore, its codebase, developed upon PyTorch, allows easy and efficient modifications to realize new tasks. 

For this work, we partially relied on contextual string embeddings (FlairEmbeddings) \cite{akbik-etal-2018-contextual} and pooled contextualized embeddings (Pool) \cite{akbik-etal-2019-pooled}. Flair embeddings are a type of word embeddings, which merge the best attributes of three embeddings (Word embeddings, Character-level features, Contextualized word embeddings). Those embeddings represent words as a sequence of characters and are contextualized by their surrounding text, making the same words have a different type of embedding depending on its context. Flair also offers the opportunity to fine-tune this kind of embeddings on unlabeled datasets for a specific domain. Pool embeddings resolve the problem of representing rare occurrences of words that might carry more than one meaning in a given text. A pooling operation is applied to all contextualized instances of a word to generate a global word representation that encodes all the gathered features into a new one.

\begin{table*}[ht!]
\centering
\caption{Part-of-speech Tagging: Average accuracy with standard deviation (in brackets) over 5-fold cross validation}
\begin{tabular}{lcccccc}
\toprule
 & \textbf{Def. Word} & \textbf{Custom Word} & \textbf{Def. Flair} & \textbf{Custom Flair} & \textbf{Def. Word+Flair} & \textbf{Cust. Word+Flair}\\ 
\midrule
\textbf{only \textit{Nephro\_Gold}} & 65.53 (0.66)  & 77.65 (0.95)    & 81.41 (0.75)    & 80.80 (0.47)    & 81.54 (0.69)  & 81.84 (1.37)\\ 
\textbf{\textit{Nephro\_Gold} \& \textit{HDT}}    & 97.07 (0.02)  & 97.29 (0.02)    & 98.47 (0.01)    & 98.37 (0.01)    & \textbf{98.57 (0.02)}  & 97.96 (0.01)\\ \midrule
\textbf{Model Size}  & 1.3 GB & 4.7 GB   & 248.7 MB & 248.7 MB & 1.6 GB & 5.0 GB\\ 
\bottomrule
\end{tabular}

\label{POS_Tagging_Results}
\end{table*}

\begin{table*}[ht!]
\centering
\caption{Concept Detection: Average micro F1 score with standard deviation (in brackets) over 5-fold cross validation}
\begin{tabular}{lccccccc}
\toprule
  & & \multicolumn{3}{c}{\textbf{Strict}} &  \multicolumn{3}{c}{\textbf{Lenient}}  \\
  \midrule
& \textbf{Model Size} & \textbf{Prec.} & \textbf{Rec. } & \textbf{F1} & \textbf{Prec.} & \textbf{Rec.} & \textbf{F1}  \\ 
\midrule
Default Word Embeddings                                & 1.3 GB& 61.20 & 57.11 & 58.97 (0.44) &  72.4 & 67.4 & 69.6 (0.54) \\ 
Custom Word Embeddings                             & 4.7 GB&  73.82 & 71.19 & 72.47 (0.61) &  83.2 & 79.4 & 81.4 (0.54) \\ 
Default Flair Embeddings                            & 248.6 MB&  74.82 & 73.85 & 74.33 (0.30) & 83.0 & 81.75 & 82.25 (0.50)  \\ 
Custom Flair Embeddings                            & 248.6 MB& 75.36 & 75.45 & 75.40 (0.46) &  83.75 & 83.25 & 83.25 (0.50)   \\ 

Default Word + Def Flair Embeddings                          & 1.6 GB &  73.20 & 72.65 & 72.92 (1.17) & 81.8 & 80.8 & 81.6 (0.54)   \\ 
Custom Word + Custom Flair Embeddings                               & 5.0 GB &  75.45 & 75.09 & 75.26 (0.56)  & 84.2 & 83.4 & 83.8 (0.83)    \\ 


Cust. Word+Cust. Flair+Cust. Pool Embeddings & 5.9 GB& 76.12 & 75.94 & \textbf{76.02} (0.64) & 84.8 & 84.2 & \textbf{84.4} (0.54)  \\
\bottomrule
\end{tabular}

\label{Concept_Detection_Results}
\end{table*}

\textit{Part-of-speech Tagging:} For the POS tagging we used a BiLSTM-CRF implementation of Flair. Overall we explored different setups with different embeddings and their combination. As the Nephro\_Gold dataset is rather small, we also ran experiments with a training and development dataset extended by Hamburg Dependency Treebank. 

\textit{Concept Detection:} For the concept detection we again relied on a BiLSTM-CRF model implemented in Flair. Similar to POS tagging, we examined a range of different embeddings and their combination. 



\subsection*{Relation Extraction}
For relation extraction, we re-implemented a CNN-based relation classifier, as proposed by Nguyen and Grishman (2015) \cite{nguyen2015relation}. 
We used word- and positional embeddings to represent the meaning and relative position of each token. Short sentences and reduced language might provide insufficient context. Thus, we also added embeddings to provide the model with concept information about the two relation arguments.


\section*{Experiments}\label{experiments}


In the following, we present our results in evaluating the models introduced in the previous section. To merge the overlapping documents, we prioritized the number of documents each student annotated (more documents, higher priority) to achieve a better consistency across the dataset. 
Then, we used JCoRe \cite{Hellrich2015} for tokenization and sentence splitting, and removed annotations crossing sentence boundaries. We also removed nested annotations and retained only the longest span. In cases where tokens were assigned multiple concept annotations, we favored the one with the higher occurrence.

We used accuracy to report results for POS tagging, and precision, recall, and micro-averaged F1 score for concept detection and relation extraction. The concept detection was evaluated by using strict and lenient matching. Strict matching considers a predicted concept to be a true positive only if its offsets exactly match the ground truth, whereas for lenient matching it is sufficient for tags to overlap, similarly as defined in Henry et al. \cite{henry20202018}.

We evaluated all models using a 5-fold cross validation with a training, development, and test split of 75/10/15. The reported results only include part-of-speech tagging, concept detection, and relation extraction, because dependency tree parsing and attribute (negation) detection results were partly already presented in previous work and tools made available (see Kara et al.~\cite{Kara:2018} and Cotik et al.~\cite{cotik-etal-2016-negation}).

\subsection*{Part-of-speech Tagging}

Table~\ref{POS_Tagging_Results} presents the results of part-of-speech tagging. The upper line shows the performance of the POS tagger using only \textit{Nephro\_Gold}, and the line below the model including the extended training and development dataset using HDT. In addition we can see the performance of the standard word and Flair embeddings, also in combination, as well as the performance of the custom embeddings. The size of each model is presented in line at the bottom. 

The table shows that combining \textit{Nephro\_Gold} with the much larger HDT data, all different setups show a strong increase of accuracy. The table also shows that custom word embeddings always outperform the default word embeddings, although the performance gain is much smaller in case of including HDT. Moreover, Flair embeddings show in all cases a boost over their word embedding equivalent. Interestingly default Flair embeddings outperform the (fine-tuned) custom Flair embeddings. Finally we can see that the combination of word and Flair embeddings tend to outperform their single equivalent - not in case of using HDT data and custom setup. However, the best setup can be achieved with HDT and both, default word and Flair embeddings. 

The presented results indicate the custom embeddings are not necessary for the POS use case. In addition to that, regarding the model size, the default Flair embedding using HDT is our favored setup, as the performance is only 0.1 below the best performing system, but the model is much smaller.

\begin{table*}[t!]
\centering
\caption{Relation Extraction: Average micro F1 score with standard deviation (in brackets) over 5-fold cross validation}
\begin{tabular}{lcccc}
\toprule
& \textbf{Model Size} &\textbf{ Prec.} & \textbf{Rec.}  & \textbf{F1}   \\ 
\midrule
Def. Word Embeddings + Relative Offsets  (Nygen et al. \cite{nguyen2015relation}) &  2.9 GB & 63.0 & 74.0 & 68.0 (0.008)\\ 

Custom Word Embeddings + Relative Offsets & 4.7 GB& 61.0 & 76.0 & 68.0 (0.007) \\
Def. Word + Concept Embeddings + Relative Offsets & 2.9 GB & 80.0 & 87.0 & 83.0 (0.01) \\ 

Custom Word + Concept Embeddings + Relative Offsets  & 4.7 GB & 79.0 & 89.0 &  \textbf{84.0} (0.003)\\ 
\bottomrule
\end{tabular}
\label{Relation_Extraction_Results}
\end{table*}

\begin{table}[ht!]
\centering
\caption{Fine-grained lenient score of (first run) concepts using custom Flair embeddings model, according to precision, recall and F1, sorted by frequency (\#). Results are presented in comparison to IAA (micro avg. F1).}
\begin{tabular}{lccccc}
\toprule
\textbf{Concept Name} & \textbf{Prec.} & \textbf{Recall} & \textbf{F1} & \textbf{\#} & \textbf{IAA}\\ 
\midrule
\textbf{Micro F1-Score} & 84.0 & 82.0 & 83.0 & &\\
\textbf{Macro F1-Score} & 81.0 & 77.0 & 79.0 & &\\
\midrule
Medical condition & 88.0 & 92.0 & 90.0 &  8953 & 87.39\\
Measurement & 77.0 & 83.0 & 80.0 &  5429 & 62.13\\
Body part & 82.0 & 85.0 & 83.0 &  3410 & 74.20\\
Treatment & 81.0 & 82.0 & 82.0 &  4379 & 79.24\\
DiagLab Procedure & 87.0 & 65.0 & 75.0 &  3209 & 66.54\\
State of health & 94.0 & 87.0 & 90.0 &  4025 & 83.32\\
Process & 86.0 & 67.0 & 76.0 &  2716 & 79.41\\
Medication & 89.0 & 90.0 & 90.0 &  3169 & 93.83\\
Time information & 89.0 & 73.0 & 80.0 &  3103 & 46.75\\
Local specification & 84.0 & 79.0 & 82.0 &  1716 & 63.83\\
Biological chemistry & 60.0 & 94.0 & 73.0 &  1363 & 71.13\\
Biological parameter & 68.0 & 66.0 & 67.0 &  966 & 60.22\\
Dosing & 93.0 & 85.0 & 88.0 &  1203 & 74.78\\
Person & 91.0 & 97.0 & 94.0 &  1265 & 85.26\\
Medical specification & 40.0 & 31.0 & 35.0 &  850 & 38.68\\
Medical device & 74.0 & 55.00 & 63.0 &  370 & 89.98\\
Body Fluid & 91.0 & 78.0 & 84.0 &  164 & 70.09\\
\bottomrule
\end{tabular}

\label{Fine_grained_Default_Flair_Results}
\end{table}

\begin{table}[ht!]
\centering
\caption{Fine-grained score (first run) of relations using ``custom word + concept embedding +
 relative offset'' model, according to precision, recall and F1, sorted by frequency (\#). Results are presented in comparison to IAA (micro avg. F1).}
\begin{tabular}{lccccc}
\toprule
\textbf{Relation Name} & \textbf{Prec.} & \textbf{Recall} & \textbf{F1} & \textbf{\#} & \textbf{IAA}\\ 
\midrule
\textbf{Micro F1-Score} & 0.77 & 0.92 & 0.84 & &\\ 
\textbf{Macro F1-Score} & 0.74 & 0.91 & 0.82 & &\\ 
\midrule
rel-has-measure  &  0.77  &  0.95  &  0.85  &3810& 62.15\\
rel-hasState  &  0.83  &  0.92  &  0.87 &2860& 76.58\\
rel-has-time-info  &  0.75  &  0.93  &  0.83  &2302& 41.24\\
rel-is-located  &  0.75  &  0.84  &  0.79  &2160& 56.38\\
rel-involves  &  0.86  &  0.95  &  0.90  &2015& 85.79\\
rel-shows  &  0.50  &  0.77  &  0.61  &1192& 65.87\\
rel-has-dosing  &  0.85  &  0.97  &  0.90  &1156& 84.97\\
rel-is-specified  &  0.78  &  0.99  &  0.87 &628& 39.19\\
rel-examines  &  0.60  &  0.89  &  0.72 &381& 57.70\\
\bottomrule
\end{tabular}

\label{Fine_grained_Relation_Results}
\end{table}

\subsection*{Concept Detection}

The results of the concept detection are presented in Table \ref{Concept_Detection_Results} and show the average F1 micro scores using a cross validation with the different setups. The evaluation was carried out using strict and lenient matching as described above.

The results show that in all cases the lenient score achieves a better performance, which is no surprise as we leave more flexibility. Moreover, we can see that the custom word embeddings outperform the default fastText emebddings. The Flair embeddings on the other hand outperform the specialized custom embeddings, and at the same time, the size of the model is below 250 MB, while the pure word embedding approaches are always above at least one GB. The overall best performing model uses a combination of word, Flair and Pool embeddings, unfortunately resulting in a model with the largest size of nearly 6GB. 

Table \ref{Fine_grained_Default_Flair_Results} shows the results of the custom Flair embeddings model on concept level. The table also shows the overall (including training and development) frequency of each concept in the dataset. As we can see, the distribution of the concepts is very unbalanced. Often multi-class approaches have problems dealing with unbalanced data. In our case the classifier can deal relatively well with the situation. Note, we also trained single classifiers for each concept, which led to marginal improvements in most cases. On the other hand this approach is not feasible for a real use-case, as each model needs to be loaded into the working memory, for only a slight overall improvement.

Moreover, Table \ref{Concept_Detection_Results} presents the micro avg. F1 IAA scores of the annotators. In most cases the IAA is below the score of the classifier, only in some cases, such as Medication, the IAA score is above. In the case of Medical Specification for instance the IAA is very low, this is also represented in the results of the machine learning model. 




\subsection*{Relation Extraction}

Table~\ref{Relation_Extraction_Results} shows the relation extraction results. The baseline model with default word embeddings (Nygen et al.) yields an F1 score of $68.0$. The usage of the custom word embeddings does lead to any improvements but increases model size by more than 3 GB. Supplementing the two models with concept embeddings considerably increases performance to $83.0$ and $84.0$ F1 for default and custom word embeddings, respectively. This supports our hypothesis that concept information is beneficial to relation extraction from clinical text, as the context lacks important linguistic information.

Table~\ref{Fine_grained_Relation_Results} shows the detailed results (first run) of the \textit{default word + concept embeddings + relative offsets} model, including the overall frequency of the different relations. Similarly to the concepts, the distribution of the relations is unbalanced. However, all relations can be detected very well, often with an F1 above 0.8.

Moreover, Table~\ref{Fine_grained_Relation_Results} presents the micro avg. F1 IAA scores of the annotators. Similarly as in Table \ref{Concept_Detection_Results}, the IAA tends to be below the results of the machine learning model. Notably, while the performance of the relation is-specified achieves quite good results with an F1 score of 87, the IAA is only about 39. It is likely that the disagreements between the annotators regarding Medical-Specification have a strong influence on that. Conversely, more than 50\% of the data was annotated by one particular annotator. This might have a strong influence on the overall annotations, and the model was probably able to learn this annotation style very well, which is reflected in the performance of that relation.

\subsection*{Medical Text Processing Workbench}

Given the previous experiments we select the best models according to score and model size. Generally we prefer a small sized model with a score slightly below the best performing system. Moreover, as the evaluation was carried out within a 5-fold cross validation, resulting in five different models, we always pick the first model for the workbench. Overall the following models will be used: POS Tagger (Default Flair, run 1), concept detection (Custom Flair, run 1) and relation extraction (Default Word+Concept+Relative Offset, run 1).

The description of the workbench, to run the models out of the box, can be found on \url{http://biomedical.dfki.de}. The models themselves can be downloaded here: \url{https://github.com/DFKI-NLP/mEx-Docker-Deployment}. Note, in order to use the models, a user agreement must be signed first.

\subsection*{Testing Concept Detection on additional Datasets}

The experiments above have been conducted on clinical text of the nephrology domain. All documents have been written within one hospital. Due to the restricted topic and due to the limited number of authors writing those reports, the data might be very homogeneous. Therefore models trained on this data might be not suitable to be tested on other biomedical/clinical texts in German. 

In order to explore this, we carried out a small proof of concept. To do so, we tested our concept detection on two additional biomedical text datasets in German, namely (a) GGPONC \cite{borchert2020ggponc}, a dataset of clinical practice guidelines and (b) a set of posts published in a German health forum, taken from the TLC corpus \cite{seiffe2020witch}. In both cases we applied the model to 600 sentences of each dataset. The automatically generated labels were then corrected (if needed) by the main annotator of our nephrology corpus. Please note, we wanted to find out if our concept detection can work in general also on different text, given our annotation schema. So the annotator examined the given labels, but was not too strict about each single label.

Regarding micro F-Score, the IAA between classifier and annotator was 0.851 on GGPONC and 0.868 on the TLC forum data. These values are surprisingly good, also in comparison to the original concept detection results on the nephrology dataset. The fact that correcting a dataset instead of starting an annotation from scratch, might have an influence, as well as accepting `suggested labels' which would have been not annotated in other cases. 
Overall the outcomes serve as a proof of concept, rather than a solid evaluation. However, the results indicate that our models might be a good first choice to process biomedical text in German. Data can be shared upon request.

\subsection*{Discussion and Analysis}\label{analysis}

We carried out experiments with different setups on three different tasks using clinical text in German. In most cases, we observed customized word embeddings to provide better results compared to their default counterpart. This performance gain, however, comes at the expense of increased model size, particularly for POS tagging and concept detection.
Moreover, within the first two experiments we could see that the models using only Flair embeddings perform quite well and tend to have a smaller model size. 
The results show that character embedding-based approaches seem to perform well on clinical text. A reason for that might be the characteristics of some words (medication names often contain letters like `x' and `y' \cite{wick2004}) and possibly the features of the Greek and Latin origin of words (e.g. words ending with `-\textit{itis}' or `\textit{iasis}' might refer to a disease).

For applying and running such text processing models on the fly in clinical care, smaller models might be favoured, in order to be not too dependent on computational power and working memory. This means that the simple Flair embedding models would probably be the better choice for a real clinical use case. In case of relation extraction instead, we relied on an implementation which does not integrate Flair or character embeddings. Therefore the favoured models still have a size of 2.9 GB. It would make sense to switch to an efficient model which results in a smaller model size.

Moreover, we analyzed the predictions of the model. Generally, the clinical data of our corpus is heterogeneous. Even though a large range of health related problems can be described in the documents, they are all of the nephrology domain and always of the same department. Therefore, we frequently observed similar text patterns for sentences, mentioning similar medical problems, treatments or medications. Regarding POS tagging, the fact that the dataset is relatively small, and contains similar sentences (or often a similar sentence structure) might have been the reason for the good results.

The concept detection certainly also benefited from frequently re-occurring concepts. On the other hand the same words could be annotated differently, depending on context, but also depending on the annotator. This particularly made the strict matching more challenging. Analyzing the falsely predicted concepts, various cases could be found where the classifier attached a label, which was not necessarily wrong. 
Interestingly our concept detection showed very promising results, when applied to two different biomedical datasets in German.

\section*{Conclusion}

In this work, we described an annotated corpus of German nephrology reports and further created a collection of German biomedical text to train customized word embeddings for clinical text. Based on this data in combination with existing methods and tools, we created and evaluated a set of German clinical text processing models: a part-of-speech tagger, a concept detector, and a relation extractor. To provide resources for the clinical text processing community, we combined the best performing models into a medical information extraction workbench, which we made publicly available for free use.





\begin{backmatter}

\section*{Acknowledgements}
N/A

\section*{Funding}
This research was supported by the German Research Foundation (Deutsche Forschungsgemeinschaft, DFG) through the project KEEPHA (442445488), by the German Federal Ministry of Education and Research (BMBF) through the projects BIFOLD (01IS18025E), and the European Union's Horizon 2020 research and innovation program under grant agreement No 780495 (BigMedilytics). MM and WD are participants in the BIH Charité Digital Clinician Scientist Program funded by the Charité – Universitätsmedizin Berlin, and the Berlin Institute of Health at Charité (BIH).

\section*{Abbreviations}
NLP - Natural Language Processing;
NER - Named Entity Recognition;
RE - Relation Extraction;
POS - Part-of-Speech Tagging; 
HDT - Hamburg Dependency Treebank;
IAA - Inter-Annotator Agreement;


\section*{Availability of data and materials}
The description of the mEx workbench can be found on http://biomedical.dfki.de. The mEx models themselves can be downloaded on https://github.com/DFKI-NLP/mEx-Docker-Deployment. 

\section*{Ethics approval and consent to participate}
N/A

\section*{Competing interests}
The authors declare that they have no competing interests.

\section*{Consent for publication}
N/A

\section*{Authors' contributions}
All co-authors are justifiably credited with authorship, according to the authorship criteria. In detail: RR- coordination, planing, development of annotation schema, writing, analysis; LR- development of annotation schema, annotations, evaluation, writing corpus section; AA- technical development, writing technical section; SM- critical revision of manuscript; OM- annotations, writing corpus section; MM- annotations, writing corpus section; CA- technical development, writing technical section; DS- technical support, data preparation; FH- development of annotation schema; MN- discussions, editing, revision of the manuscript; WD- discussions, editing, revision of the manuscript; KB- planing, critical revision of manuscript. All authors read and approved the final manuscript.



\bibliographystyle{bmc-mathphys} 
\bibliography{bmc_article}      


\begin{thebibliography}{62}
\ifx \bisbn   \undefined \def \bisbn  #1{ISBN #1}\fi
\ifx \binits  \undefined \def \binits#1{#1}\fi
\ifx \bauthor  \undefined \def \bauthor#1{#1}\fi
\ifx \batitle  \undefined \def \batitle#1{#1}\fi
\ifx \bjtitle  \undefined \def \bjtitle#1{#1}\fi
\ifx \bvolume  \undefined \def \bvolume#1{\textbf{#1}}\fi
\ifx \byear  \undefined \def \byear#1{#1}\fi
\ifx \bissue  \undefined \def \bissue#1{#1}\fi
\ifx \bfpage  \undefined \def \bfpage#1{#1}\fi
\ifx \blpage  \undefined \def \blpage #1{#1}\fi
\ifx \burl  \undefined \def \burl#1{\textsf{#1}}\fi
\ifx \doiurl  \undefined \def \doiurl#1{\textsf{#1}}\fi
\ifx \betal  \undefined \def \betal{\textit{et al.}}\fi
\ifx \binstitute  \undefined \def \binstitute#1{#1}\fi
\ifx \binstitutionaled  \undefined \def \binstitutionaled#1{#1}\fi
\ifx \bctitle  \undefined \def \bctitle#1{#1}\fi
\ifx \beditor  \undefined \def \beditor#1{#1}\fi
\ifx \bpublisher  \undefined \def \bpublisher#1{#1}\fi
\ifx \bbtitle  \undefined \def \bbtitle#1{#1}\fi
\ifx \bedition  \undefined \def \bedition#1{#1}\fi
\ifx \bseriesno  \undefined \def \bseriesno#1{#1}\fi
\ifx \blocation  \undefined \def \blocation#1{#1}\fi
\ifx \bsertitle  \undefined \def \bsertitle#1{#1}\fi
\ifx \bsnm \undefined \def \bsnm#1{#1}\fi
\ifx \bsuffix \undefined \def \bsuffix#1{#1}\fi
\ifx \bparticle \undefined \def \bparticle#1{#1}\fi
\ifx \barticle \undefined \def \barticle#1{#1}\fi
\ifx \bconfdate \undefined \def \bconfdate #1{#1}\fi
\ifx \botherref \undefined \def \botherref #1{#1}\fi
\ifx \url \undefined \def \url#1{\textsf{#1}}\fi
\ifx \bchapter \undefined \def \bchapter#1{#1}\fi
\ifx \bbook \undefined \def \bbook#1{#1}\fi
\ifx \bcomment \undefined \def \bcomment#1{#1}\fi
\ifx \oauthor \undefined \def \oauthor#1{#1}\fi
\ifx \citeauthoryear \undefined \def \citeauthoryear#1{#1}\fi
\ifx \endbibitem  \undefined \def \endbibitem {}\fi
\ifx \bconflocation  \undefined \def \bconflocation#1{#1}\fi
\ifx \arxivurl  \undefined \def \arxivurl#1{\textsf{#1}}\fi
\csname PreBibitemsHook\endcsname

\bibitem{savova2010mayo}
\begin{barticle}
\bauthor{\bsnm{Savova}, \binits{G.K.}},
\bauthor{\bsnm{Masanz}, \binits{J.J.}},
\bauthor{\bsnm{Ogren}, \binits{P.V.}},
\bauthor{\bsnm{Zheng}, \binits{J.}},
\bauthor{\bsnm{Sohn}, \binits{S.}},
\bauthor{\bsnm{Kipper-Schuler}, \binits{K.C.}},
\bauthor{\bsnm{Chute}, \binits{C.G.}}:
\batitle{Mayo clinical text analysis and knowledge extraction system (ctakes):
  architecture, component evaluation and applications}.
\bjtitle{Journal of the American Medical Informatics Association}
\bvolume{17}(\bissue{5}),
\bfpage{507}--\blpage{513}
(\byear{2010})
\end{barticle}
\endbibitem

\bibitem{aronson2010overview}
\begin{barticle}
\bauthor{\bsnm{Aronson}, \binits{A.R.}},
\bauthor{\bsnm{Lang}, \binits{F.-M.}}:
\batitle{An overview of metamap: historical perspective and recent advances}.
\bjtitle{Journal of the American Medical Informatics Association}
\bvolume{17}(\bissue{3}),
\bfpage{229}--\blpage{236}
(\byear{2010})
\end{barticle}
\endbibitem

\bibitem{friedman1994general}
\begin{barticle}
\bauthor{\bsnm{Friedman}, \binits{C.}},
\bauthor{\bsnm{Alderson}, \binits{P.O.}},
\bauthor{\bsnm{Austin}, \binits{J.H.}},
\bauthor{\bsnm{Cimino}, \binits{J.J.}},
\bauthor{\bsnm{Johnson}, \binits{S.B.}}:
\batitle{A general natural-language text processor for clinical radiology}.
\bjtitle{Journal of the American Medical Informatics Association}
\bvolume{1}(\bissue{2}),
\bfpage{161}--\blpage{174}
(\byear{1994})
\end{barticle}
\endbibitem

\bibitem{xu2010medex}
\begin{barticle}
\bauthor{\bsnm{Xu}, \binits{H.}},
\bauthor{\bsnm{Stenner}, \binits{S.P.}},
\bauthor{\bsnm{Doan}, \binits{S.}},
\bauthor{\bsnm{Johnson}, \binits{K.B.}},
\bauthor{\bsnm{Waitman}, \binits{L.R.}},
\bauthor{\bsnm{Denny}, \binits{J.C.}}:
\batitle{Medex: a medication information extraction system for clinical
  narratives}.
\bjtitle{Journal of the American Medical Informatics Association}
\bvolume{17}(\bissue{1}),
\bfpage{19}--\blpage{24}
(\byear{2010})
\end{barticle}
\endbibitem

\bibitem{sohn2013comprehensive}
\begin{barticle}
\bauthor{\bsnm{Sohn}, \binits{S.}},
\bauthor{\bsnm{Wagholikar}, \binits{K.B.}},
\bauthor{\bsnm{Li}, \binits{D.}},
\bauthor{\bsnm{Jonnalagadda}, \binits{S.R.}},
\bauthor{\bsnm{Tao}, \binits{C.}},
\bauthor{\bsnm{Komandur~Elayavilli}, \binits{R.}},
\bauthor{\bsnm{Liu}, \binits{H.}}:
\batitle{Comprehensive temporal information detection from clinical text:
  medical events, time, and tlink identification}.
\bjtitle{Journal of the American Medical Informatics Association}
\bvolume{20}(\bissue{5}),
\bfpage{836}--\blpage{842}
(\byear{2013})
\end{barticle}
\endbibitem

\bibitem{WANG201834}
\begin{barticle}
\bauthor{\bsnm{Wang}, \binits{Y.}},
\bauthor{\bsnm{Wang}, \binits{L.}},
\bauthor{\bsnm{Rastegar-Mojarad}, \binits{M.}},
\bauthor{\bsnm{Moon}, \binits{S.}},
\bauthor{\bsnm{Shen}, \binits{F.}},
\bauthor{\bsnm{Afzal}, \binits{N.}},
\bauthor{\bsnm{Liu}, \binits{S.}},
\bauthor{\bsnm{Zeng}, \binits{Y.}},
\bauthor{\bsnm{Mehrabi}, \binits{S.}},
\bauthor{\bsnm{Sohn}, \binits{S.}},
\bauthor{\bsnm{Liu}, \binits{H.}}:
\batitle{Clinical information extraction applications: A literature review}.
\bjtitle{Journal of Biomedical Informatics}
\bvolume{77},
\bfpage{34}--\blpage{49}
(\byear{2018})
\end{barticle}
\endbibitem

\bibitem{uzuner2010extracting}
\begin{barticle}
\bauthor{\bsnm{Uzuner}, \binits{{\"O}.}},
\bauthor{\bsnm{Solti}, \binits{I.}},
\bauthor{\bsnm{Cadag}, \binits{E.}}:
\batitle{Extracting medication information from clinical text}.
\bjtitle{Journal of the American Medical Informatics Association}
\bvolume{17}(\bissue{5}),
\bfpage{514}--\blpage{518}
(\byear{2010})
\end{barticle}
\endbibitem

\bibitem{stubbs2015identifying}
\begin{barticle}
\bauthor{\bsnm{Stubbs}, \binits{A.}},
\bauthor{\bsnm{Kotfila}, \binits{C.}},
\bauthor{\bsnm{Xu}, \binits{H.}},
\bauthor{\bsnm{Uzuner}, \binits{{\"O}.}}:
\batitle{Identifying risk factors for heart disease over time: Overview of 2014
  i2b2/uthealth shared task track 2}.
\bjtitle{Journal of biomedical informatics}
\bvolume{58},
\bfpage{67}--\blpage{77}
(\byear{2015})
\end{barticle}
\endbibitem

\bibitem{suominen2013overview}
\begin{bchapter}
\bauthor{\bsnm{Suominen}, \binits{H.}},
\bauthor{\bsnm{Salanter{\"a}}, \binits{S.}},
\bauthor{\bsnm{Velupillai}, \binits{S.}},
\bauthor{\bsnm{Chapman}, \binits{W.W.}},
\bauthor{\bsnm{Savova}, \binits{G.}},
\bauthor{\bsnm{Elhadad}, \binits{N.}},
\bauthor{\bsnm{Pradhan}, \binits{S.}},
\bauthor{\bsnm{South}, \binits{B.R.}},
\bauthor{\bsnm{Mowery}, \binits{D.L.}},
\bauthor{\bsnm{Jones}, \binits{G.J.}}, \betal:
\bctitle{Overview of the share/clef ehealth evaluation lab 2013}.
In: \bbtitle{International Conference of the Cross-Language Evaluation Forum
  for European Languages},
pp. \bfpage{212}--\blpage{231}
(\byear{2013}).
\bcomment{Springer}
\end{bchapter}
\endbibitem

\bibitem{kelly2014overview}
\begin{bchapter}
\bauthor{\bsnm{Kelly}, \binits{L.}},
\bauthor{\bsnm{Goeuriot}, \binits{L.}},
\bauthor{\bsnm{Suominen}, \binits{H.}},
\bauthor{\bsnm{Schreck}, \binits{T.}},
\bauthor{\bsnm{Leroy}, \binits{G.}},
\bauthor{\bsnm{Mowery}, \binits{D.L.}},
\bauthor{\bsnm{Velupillai}, \binits{S.}},
\bauthor{\bsnm{Chapman}, \binits{W.W.}},
\bauthor{\bsnm{Martinez}, \binits{D.}},
\bauthor{\bsnm{Zuccon}, \binits{G.}}, \betal:
\bctitle{Overview of the share/clef ehealth evaluation lab 2014}.
In: \bbtitle{International Conference of the Cross-Language Evaluation Forum
  for European Languages},
pp. \bfpage{172}--\blpage{191}
(\byear{2014}).
\bcomment{Springer}
\end{bchapter}
\endbibitem

\bibitem{pradhan2014semeval}
\begin{bchapter}
\bauthor{\bsnm{Pradhan}, \binits{S.}},
\bauthor{\bsnm{Elhadad}, \binits{N.}},
\bauthor{\bsnm{Chapman}, \binits{W.}},
\bauthor{\bsnm{Manandhar}, \binits{S.}},
\bauthor{\bsnm{Savova}, \binits{G.}}:
\bctitle{Semeval-2014 task 7: Analysis of clinical text}.
In: \bbtitle{Proceedings of the 8th International Workshop on Semantic
  Evaluation (SemEval 2014)},
pp. \bfpage{54}--\blpage{62}
(\byear{2014})
\end{bchapter}
\endbibitem

\bibitem{bethard2016semeval}
\begin{bchapter}
\bauthor{\bsnm{Bethard}, \binits{S.}},
\bauthor{\bsnm{Savova}, \binits{G.}},
\bauthor{\bsnm{Chen}, \binits{W.-T.}},
\bauthor{\bsnm{Derczynski}, \binits{L.}},
\bauthor{\bsnm{Pustejovsky}, \binits{J.}},
\bauthor{\bsnm{Verhagen}, \binits{M.}}:
\bctitle{Semeval-2016 task 12: Clinical tempeval}.
In: \bbtitle{Proceedings of the 10th International Workshop on Semantic
  Evaluation (SemEval-2016)},
pp. \bfpage{1052}--\blpage{1062}
(\byear{2016})
\end{bchapter}
\endbibitem

\bibitem{neveol2018clef}
\begin{bchapter}
\bauthor{\bsnm{N{\'e}v{\'e}ol}, \binits{A.}},
\bauthor{\bsnm{Robert}, \binits{A.}},
\bauthor{\bsnm{Grippo}, \binits{F.}},
\bauthor{\bsnm{Morgand}, \binits{C.}},
\bauthor{\bsnm{Orsi}, \binits{C.}},
\bauthor{\bsnm{Pelikan}, \binits{L.}},
\bauthor{\bsnm{Ramadier}, \binits{L.}},
\bauthor{\bsnm{Rey}, \binits{G.}},
\bauthor{\bsnm{Zweigenbaum}, \binits{P.}}:
\bctitle{Clef ehealth 2018 multilingual information extraction task overview:
  Icd10 coding of death certificates in french, hungarian and italian.}
In: \bbtitle{CLEF (Working Notes)}
(\byear{2018})
\end{bchapter}
\endbibitem

\bibitem{jimenez2019neges}
\begin{barticle}
\bauthor{\bsnm{Jim{\'e}nez-Zafra}, \binits{S.M.}},
\bauthor{\bsnm{D{\'\i}az}, \binits{N.P.C.}},
\bauthor{\bsnm{Morante}, \binits{R.}},
\bauthor{\bsnm{Mart{\'\i}n-Valdivia}, \binits{M.T.}}:
\batitle{Neges 2018: Workshop on negation in spanish}.
\bjtitle{Procesamiento del Lenguaje Natural}
\bvolume{62},
\bfpage{21}--\blpage{28}
(\byear{2019})
\end{barticle}
\endbibitem

\bibitem{neveol2018clinical}
\begin{barticle}
\bauthor{\bsnm{N{\'e}v{\'e}ol}, \binits{A.}},
\bauthor{\bsnm{Dalianis}, \binits{H.}},
\bauthor{\bsnm{Velupillai}, \binits{S.}},
\bauthor{\bsnm{Savova}, \binits{G.}},
\bauthor{\bsnm{Zweigenbaum}, \binits{P.}}:
\batitle{Clinical natural language processing in languages other than english:
  opportunities and challenges}.
\bjtitle{Journal of biomedical semantics}
\bvolume{9}(\bissue{1}),
\bfpage{12}
(\byear{2018})
\end{barticle}
\endbibitem

\bibitem{dorendahl2019overview}
\begin{botherref}
\oauthor{\bsnm{D{\"o}rendahl}, \binits{A.}},
\oauthor{\bsnm{Leich}, \binits{N.}},
\oauthor{\bsnm{Hummel}, \binits{B.}},
\oauthor{\bsnm{Sch{\"o}nfelder}, \binits{G.}},
\oauthor{\bsnm{Grune}, \binits{B.}}:
Overview of the clef ehealth 2019 multilingual information extraction
(2019)
\end{botherref}
\endbibitem

\bibitem{borchert2020ggponc}
\begin{bchapter}
\bauthor{\bsnm{Borchert}, \binits{F.}},
\bauthor{\bsnm{Lohr}, \binits{C.}},
\bauthor{\bsnm{Modersohn}, \binits{L.}},
\bauthor{\bsnm{Langer}, \binits{T.}},
\bauthor{\bsnm{Follmann}, \binits{M.}},
\bauthor{\bsnm{Sachs}, \binits{J.P.}},
\bauthor{\bsnm{Hahn}, \binits{U.}},
\bauthor{\bsnm{Schapranow}, \binits{M.-P.}}:
\bctitle{Ggponc: A corpus of german medical text with rich metadata based on
  clinical practice guidelines}.
In: \bbtitle{Proceedings of the 11th International Workshop on Health Text
  Mining and Information Analysis},
pp. \bfpage{38}--\blpage{48}
(\byear{2020})
\end{bchapter}
\endbibitem

\bibitem{seiffe2020witch}
\begin{bchapter}
\bauthor{\bsnm{Seiffe}, \binits{L.}},
\bauthor{\bsnm{Marten}, \binits{O.}},
\bauthor{\bsnm{Mikhailov}, \binits{M.}},
\bauthor{\bsnm{Schmeier}, \binits{S.}},
\bauthor{\bsnm{M{\"o}ller}, \binits{S.}},
\bauthor{\bsnm{Roller}, \binits{R.}}:
\bctitle{From witch’s shot to music making bones-resources for medical laymen
  to technical language and vice versa}.
In: \bbtitle{Proceedings of the 12th Language Resources and Evaluation
  Conference},
pp. \bfpage{6185}--\blpage{6192}
(\byear{2020})
\end{bchapter}
\endbibitem

\bibitem{lohr2018sharing}
\begin{bchapter}
\bauthor{\bsnm{Lohr}, \binits{C.}},
\bauthor{\bsnm{Buechel}, \binits{S.}},
\bauthor{\bsnm{Hahn}, \binits{U.}}:
\bctitle{Sharing copies of synthetic clinical corpora without physical
  distribution—a case study to get around iprs and privacy constraints
  featuring the german jsyncc corpus}.
In: \bbtitle{Proceedings of the Eleventh International Conference on Language
  Resources and Evaluation (LREC 2018)}
(\byear{2018})
\end{bchapter}
\endbibitem

\bibitem{Hellrich2015}
\begin{barticle}
\bauthor{\bsnm{Hellrich}, \binits{J.}},
\bauthor{\bsnm{Matthies}, \binits{F.}},
\bauthor{\bsnm{Faessler}, \binits{E.}},
\bauthor{\bsnm{Hahn}, \binits{U.}}:
\batitle{{Sharing Models and Tools for Processing German Clinical Texts}}.
\bjtitle{MIE 2015 - Digital Healthcare Empowering Europeans}
\bvolume{210},
\bfpage{734}--\blpage{738}
(\byear{2015})
\end{barticle}
\endbibitem

\bibitem{wermter2004really}
\begin{bchapter}
\bauthor{\bsnm{Wermter}, \binits{J.}},
\bauthor{\bsnm{Hahn}, \binits{U.}}:
\bctitle{Really, is medical sublanguage that different? experimental
  counter-evidence from tagging medical and newspaper corpora.}
In: \bbtitle{Medinfo},
pp. \bfpage{560}--\blpage{564}
(\byear{2004})
\end{bchapter}
\endbibitem

\bibitem{CHAPMAN2001301}
\begin{barticle}
\bauthor{\bsnm{Chapman}, \binits{W.W.}},
\bauthor{\bsnm{Bridewell}, \binits{W.}},
\bauthor{\bsnm{Hanbury}, \binits{P.}},
\bauthor{\bsnm{Cooper}, \binits{G.F.}},
\bauthor{\bsnm{Buchanan}, \binits{B.G.}}:
\batitle{A simple algorithm for identifying negated findings and diseases in
  discharge summaries}.
\bjtitle{Journal of Biomedical Informatics}
\bvolume{34}(\bissue{5}),
\bfpage{301}--\blpage{310}
(\byear{2001})
\end{barticle}
\endbibitem

\bibitem{chapman2013extending}
\begin{barticle}
\bauthor{\bsnm{Chapman}, \binits{W.W.}},
\bauthor{\bsnm{Hilert}, \binits{D.}},
\bauthor{\bsnm{Velupillai}, \binits{S.}},
\bauthor{\bsnm{Kvist}, \binits{M.}},
\bauthor{\bsnm{Skeppstedt}, \binits{M.}},
\bauthor{\bsnm{Chapman}, \binits{B.E.}},
\bauthor{\bsnm{Conway}, \binits{M.}},
\bauthor{\bsnm{Tharp}, \binits{M.}},
\bauthor{\bsnm{Mowery}, \binits{D.L.}},
\bauthor{\bsnm{Deleger}, \binits{L.}}:
\batitle{Extending the negex lexicon for multiple languages}.
\bjtitle{Studies in health technology and informatics}
\bvolume{192},
\bfpage{677}
(\byear{2013})
\end{barticle}
\endbibitem

\bibitem{cotik-etal-2016-negation}
\begin{bchapter}
\bauthor{\bsnm{Cotik}, \binits{V.}},
\bauthor{\bsnm{Roller}, \binits{R.}},
\bauthor{\bsnm{Xu}, \binits{F.}},
\bauthor{\bsnm{Uszkoreit}, \binits{H.}},
\bauthor{\bsnm{Budde}, \binits{K.}},
\bauthor{\bsnm{Schmidt}, \binits{D.}}:
\bctitle{Negation detection in clinical reports written in {G}erman}.
In: \bbtitle{Proceedings of the Fifth Workshop on Building and Evaluating
  Resources for Biomedical Text Mining ({B}io{T}xt{M}2016)},
pp. \bfpage{115}--\blpage{124}.
\bpublisher{The COLING 2016 Organizing Committee},
\blocation{Osaka, Japan}
(\byear{2016})
\end{bchapter}
\endbibitem

\bibitem{Kara:2018}
\begin{bchapter}
\bauthor{\bsnm{Kara}, \binits{E.}},
\bauthor{\bsnm{Zeen}, \binits{T.}},
\bauthor{\bsnm{Gabryszak}, \binits{A.}},
\bauthor{\bsnm{Budde}, \binits{K.}},
\bauthor{\bsnm{Schmidt}, \binits{D.}},
\bauthor{\bsnm{Roller}, \binits{R.}}:
\bctitle{{A Domain-adapted Dependency Parser for German Clinical Text}}.
In: \bbtitle{Proceedings of the 14th Conference on Natural Language Processing
  (KONVENS 2018)},
\bconflocation{Vienna, Austria}
(\byear{2018})
\end{bchapter}
\endbibitem

\bibitem{oleynik2017unsupervised}
\begin{barticle}
\bauthor{\bsnm{Oleynik}, \binits{M.}},
\bauthor{\bsnm{Kreuzthaler}, \binits{M.}},
\bauthor{\bsnm{Schulz}, \binits{S.}}:
\batitle{Unsupervised abbreviation expansion in clinical narratives.}
\bjtitle{MedInfo}
\bvolume{245},
\bfpage{539}--\blpage{543}
(\byear{2017})
\end{barticle}
\endbibitem

\bibitem{lohr2021pseudonymization}
\begin{bchapter}
\bauthor{\bsnm{Lohr}, \binits{C.}},
\bauthor{\bsnm{Eder}, \binits{E.}},
\bauthor{\bsnm{Hahn}, \binits{U.}}:
\bctitle{{Pseudonymization of PHI items in German clinical reports}}.
In: \bbtitle{Public Health and Informatics}
(\byear{2021})
\end{bchapter}
\endbibitem

\bibitem{kittner2021annotation}
\begin{barticle}
\bauthor{\bsnm{Kittner}, \binits{M.}},
\bauthor{\bsnm{Lamping}, \binits{M.}},
\bauthor{\bsnm{Rieke}, \binits{D.T.}},
\bauthor{\bsnm{G{\"o}tze}, \binits{J.}},
\bauthor{\bsnm{Bajwa}, \binits{B.}},
\bauthor{\bsnm{Jelas}, \binits{I.}},
\bauthor{\bsnm{R{\"u}ter}, \binits{G.}},
\bauthor{\bsnm{Hautow}, \binits{H.}},
\bauthor{\bsnm{S{\"a}nger}, \binits{M.}},
\bauthor{\bsnm{Habibi}, \binits{M.}}, \betal:
\batitle{Annotation and initial evaluation of a large annotated german
  oncological corpus}.
\bjtitle{JAMIA open}
\bvolume{4}(\bissue{2}),
\bfpage{025}
(\byear{2021})
\end{barticle}
\endbibitem

\bibitem{frei2022gernermed}
\begin{barticle}
\bauthor{\bsnm{Frei}, \binits{J.}},
\bauthor{\bsnm{Kramer}, \binits{F.}}:
\batitle{Gernermed: An open german medical ner model}.
\bjtitle{Software Impacts}
\bvolume{11},
\bfpage{100212}
(\byear{2022})
\end{barticle}
\endbibitem

\bibitem{henry20202018}
\begin{barticle}
\bauthor{\bsnm{Henry}, \binits{S.}},
\bauthor{\bsnm{Buchan}, \binits{K.}},
\bauthor{\bsnm{Filannino}, \binits{M.}},
\bauthor{\bsnm{Stubbs}, \binits{A.}},
\bauthor{\bsnm{Uzuner}, \binits{O.}}:
\batitle{2018 n2c2 shared task on adverse drug events and medication extraction
  in electronic health records}.
\bjtitle{Journal of the American Medical Informatics Association}
\bvolume{27}(\bissue{1}),
\bfpage{3}--\blpage{12}
(\byear{2020})
\end{barticle}
\endbibitem

\bibitem{shrestha2021development}
\begin{botherref}
\oauthor{\bsnm{Shrestha}, \binits{M.}}:
Development of a language model for medical domain.
Master's thesis,
Hochschule Rhein-Waal
(2021)
\end{botherref}
\endbibitem

\bibitem{HFGermanMedBERT}
\begin{botherref}
German-MedBERT on Hugging Face.
\url{https://huggingface.co/smanjil/German-MedBERT}.
Accessed: 2022-03-15
\end{botherref}
\endbibitem

\bibitem{Seuss2017}
\begin{bchapter}
\bauthor{\bsnm{Seuss}, \binits{H.}},
\bauthor{\bsnm{Dankerl}, \binits{P.}},
\bauthor{\bsnm{Ihle}, \binits{M.}},
\bauthor{\bsnm{Grandjean}, \binits{A.}},
\bauthor{\bsnm{Hammon}, \binits{R.}},
\bauthor{\bsnm{Kaestle}, \binits{N.}},
\bauthor{\bsnm{Fasching}, \binits{P.A.}},
\bauthor{\bsnm{Maier}, \binits{C.}},
\bauthor{\bsnm{Christoph}, \binits{J.}},
\bauthor{\bsnm{Sedlmayr}, \binits{M.}}, \betal:
\bctitle{Semi-automated de-identification of german content sensitive reports
  for big data analytics}.
In: \bbtitle{R{\"o}Fo-Fortschritte Auf dem Gebiet der R{\"o}ntgenstrahlen und
  der Bildgebenden Verfahren},
vol. \bseriesno{189},
pp. \bfpage{661}--\blpage{671}
(\byear{2017}).
\bcomment{{\copyright} Georg Thieme Verlag KG}
\end{bchapter}
\endbibitem

\bibitem{Beckers}
\begin{botherref}
Beckers Abkürzungslexikon Medizinischer Begriffe.
\url{https://www.medizinische-abkuerzungen.de/suche.html}.
Accessed: 2021-01-10
\end{botherref}
\endbibitem

\bibitem{Roller:2016}
\begin{bchapter}
\bauthor{\bsnm{Roller}, \binits{R.}},
\bauthor{\bsnm{Uszkoreit}, \binits{H.}},
\bauthor{\bsnm{Xu}, \binits{F.}},
\bauthor{\bsnm{Seiffe}, \binits{L.}},
\bauthor{\bsnm{Mikhailov}, \binits{M.}},
\bauthor{\bsnm{Staeck}, \binits{O.}},
\bauthor{\bsnm{Budde}, \binits{K.}},
\bauthor{\bsnm{Halleck}, \binits{F.}},
\bauthor{\bsnm{Schmidt}, \binits{D.}}:
\bctitle{A fine-grained corpus annotation schema of german nephrology records}.
In: \bbtitle{Proceedings of the Clinical Natural Language Processing Workshop
  (ClinicalNLP)},
pp. \bfpage{69}--\blpage{77}.
\bpublisher{The COLING 2016 Organizing Committee},
\blocation{Osaka, Japan}
(\byear{2016})
\end{bchapter}
\endbibitem

\bibitem{Stenetorp:2012}
\begin{bchapter}
\bauthor{\bsnm{Stenetorp}, \binits{P.}},
\bauthor{\bsnm{Pyysalo}, \binits{S.}},
\bauthor{\bsnm{Topi\'{c}}, \binits{G.}},
\bauthor{\bsnm{Ohta}, \binits{T.}},
\bauthor{\bsnm{Ananiadou}, \binits{S.}},
\bauthor{\bsnm{Tsujii}, \binits{J.}}:
\bctitle{{brat}: a web-based tool for {NLP}-assisted text annotation}.
In: \bbtitle{Proceedings of the Demonstrations Session at {EACL} 2012}.
\bpublisher{Association for Computational Linguistics},
\blocation{Avignon, France}
(\byear{2012})
\end{bchapter}
\endbibitem

\bibitem{hripcsak2005agreement}
\begin{barticle}
\bauthor{\bsnm{Hripcsak}, \binits{G.}},
\bauthor{\bsnm{Rothschild}, \binits{A.S.}}:
\batitle{Agreement, the f-measure, and reliability in information retrieval}.
\bjtitle{Journal of the American medical informatics association}
\bvolume{12}(\bissue{3}),
\bfpage{296}--\blpage{298}
(\byear{2005})
\end{barticle}
\endbibitem

\bibitem{Seiffe:2018}
\begin{botherref}
\oauthor{\bsnm{Seiffe}, \binits{L.}}:
{Linguistic Modeling for Text Analytic Tasks for German Clinical Texts}.
Master's thesis,
TU Berlin
(2018)
\end{botherref}
\endbibitem

\bibitem{schiller1999guidelines}
\begin{botherref}
\oauthor{\bsnm{Schiller}, \binits{A.}},
\oauthor{\bsnm{Teufel}, \binits{S.}},
\oauthor{\bsnm{Thielen}, \binits{C.}}:
{Guidelines f{\"u}r das Tagging deutscher Textcorpora mit STTS}.
Universit{\"a}ten Stuttgart und T{\"u}bingen
(1999)
\end{botherref}
\endbibitem

\bibitem{nivre2016}
\begin{bchapter}
\bauthor{\bsnm{Nivre}, \binits{J.}},
\bauthor{\bsnm{De~Marneffe}, \binits{M.-C.}},
\bauthor{\bsnm{Ginter}, \binits{F.}},
\bauthor{\bsnm{Goldberg}, \binits{Y.}},
\bauthor{\bsnm{Hajic}, \binits{J.}},
\bauthor{\bsnm{Manning}, \binits{C.D.}},
\bauthor{\bsnm{McDonald}, \binits{R.}},
\bauthor{\bsnm{Petrov}, \binits{S.}},
\bauthor{\bsnm{Pyysalo}, \binits{S.}},
\bauthor{\bsnm{Silveira}, \binits{N.}}, \betal:
\bctitle{Universal dependencies v1: A multilingual treebank collection}.
In: \bbtitle{Proceedings of the Tenth International Conference on Language
  Resources and Evaluation (LREC'16)},
pp. \bfpage{1659}--\blpage{1666}
(\byear{2016})
\end{bchapter}
\endbibitem

\bibitem{Foth:2014}
\begin{bchapter}
\bauthor{\bsnm{Foth}, \binits{K.A.}},
\bauthor{\bsnm{Köhn}, \binits{A.}},
\bauthor{\bsnm{Beuck}, \binits{N.}},
\bauthor{\bsnm{Menzel}, \binits{W.}}:
\bctitle{Because size does matter: The hamburg dependency treebank}.
In: \beditor{\bsnm{Chair)}, \binits{N.C.C.}},
\beditor{\bsnm{Choukri}, \binits{K.}},
\beditor{\bsnm{Declerck}, \binits{T.}},
\beditor{\bsnm{Loftsson}, \binits{H.}},
\beditor{\bsnm{Maegaard}, \binits{B.}},
\beditor{\bsnm{Mariani}, \binits{J.}},
\beditor{\bsnm{Moreno}, \binits{A.}},
\beditor{\bsnm{Odijk}, \binits{J.}},
\beditor{\bsnm{Piperidis}, \binits{S.}} (eds.)
\bbtitle{Proceedings of the Ninth International Conference on Language
  Resources and Evaluation (LREC'14)}.
\bpublisher{European Language Resources Association (ELRA)},
\blocation{Reykjavik, Iceland}
(\byear{2014})
\end{bchapter}
\endbibitem

\bibitem{Med1}
\begin{botherref}
Med1 Forum.
\url{https://med2-forum.de/}.
Accessed: 2021-01-10; Note, the Med1 forum does not exist anymore, see med2
  instead.
\end{botherref}
\endbibitem

\bibitem{DeutschesMedizinForum}
\begin{botherref}
Deutsches Medizin Forum.
\url{https://www.medizin-forum.de/}.
Accessed: 2021-01-10
\end{botherref}
\endbibitem

\bibitem{Spiegel}
\begin{botherref}
Spiegel.
\url{https://www.spiegel.de/gesundheit/}.
Accessed: 2021-01-10
\end{botherref}
\endbibitem

\bibitem{AerzteBlatt}
\begin{botherref}
Aerzte-Blatt.
\url{https://www.aerzteblatt.de/}.
Accessed: 2021-01-10
\end{botherref}
\endbibitem

\bibitem{NetDoktor}
\begin{botherref}
NetDoktor.
\url{https://www.netdoktor.de/}.
Accessed: 2021-01-10
\end{botherref}
\endbibitem

\bibitem{Onmeda}
\begin{botherref}
Onmeda.
\url{https://www.onmeda.de/}.
Accessed: 2021-01-10
\end{botherref}
\endbibitem

\bibitem{GermanPubMed}
\begin{botherref}
German PubMed Abstracts.
\url{https://pubmed.ncbi.nlm.nih.gov/}.
Accessed: 2021-01-10
\end{botherref}
\endbibitem

\bibitem{eDocTrainer}
\begin{botherref}
eDocTrainer.
\url{https://www.edoctrainer.de/}.
Accessed: 2021-01-10
\end{botherref}
\endbibitem

\bibitem{siewert2009chirurgie}
\begin{botherref}
\oauthor{\bsnm{Siewert}, \binits{J.R.}},
\oauthor{\bsnm{Stein}, \binits{H.J.}},
\oauthor{\bsnm{Allg{\"o}wer}, \binits{M.}}:
Chirurgie: mit integriertem fallquiz - 40 f{\"a}lle nach neuer ao.
Springer
(2009)
\end{botherref}
\endbibitem

\bibitem{poeck2006neurologie}
\begin{botherref}
\oauthor{\bsnm{Poeck}, \binits{K.}},
\oauthor{\bsnm{Hacke}, \binits{W.}}:
Neurologie.
Springer
(2006)
\end{botherref}
\endbibitem

\bibitem{hautmann2014urologie}
\begin{botherref}
\oauthor{\bsnm{Hautmann}, \binits{R.}},
\oauthor{\bsnm{Gschwend}, \binits{J.E.}}:
Urologie.
Springer
(2014)
\end{botherref}
\endbibitem

\bibitem{walter2016basiswissen}
\begin{bbook}
\bauthor{\bsnm{Walter}, \binits{P.}},
\bauthor{\bsnm{Plange}, \binits{N.}}:
\bbtitle{Basiswissen Augenheilkunde}.
\bpublisher{Springer}, \blocation{???}
(\byear{2017})
\end{bbook}
\endbibitem

\bibitem{lenarz2012hals}
\begin{botherref}
\oauthor{\bsnm{Lenarz}, \binits{T.}},
\oauthor{\bsnm{Boenninghaus}, \binits{H.G.}}:
Hals-nasen-ohren-heilkunde.
Springer
(2012)
\end{botherref}
\endbibitem

\bibitem{ziegenfuss2016notfallmedizin}
\begin{botherref}
\oauthor{\bsnm{Ziegenfu{\ss}}, \binits{T.}}:
Notfallmedizin.
Springer
(2016)
\end{botherref}
\endbibitem

\bibitem{goebeler2017basiswissen}
\begin{botherref}
\oauthor{\bsnm{Goebeler}, \binits{M.}},
\oauthor{\bsnm{Hamm}, \binits{H.}}:
Basiswissen dermatologie.
Springer
(2017)
\end{botherref}
\endbibitem

\bibitem{arolt2011basiswissen}
\begin{botherref}
\oauthor{\bsnm{Arolt}, \binits{V.}},
\oauthor{\bsnm{Reimer}, \binits{C.}},
\oauthor{\bsnm{Dilling}, \binits{H.}}:
Basiswissen psychiatrie und psychotherapie.
Springer
(2011)
\end{botherref}
\endbibitem

\bibitem{bojanowski2016enriching}
\begin{barticle}
\bauthor{\bsnm{Bojanowski}, \binits{P.}},
\bauthor{\bsnm{Grave}, \binits{E.}},
\bauthor{\bsnm{Joulin}, \binits{A.}},
\bauthor{\bsnm{Mikolov}, \binits{T.}}:
\batitle{Enriching word vectors with subword information}.
\bjtitle{Transactions of the association for computational linguistics}
\bvolume{5},
\bfpage{135}--\blpage{146}
(\byear{2017})
\end{barticle}
\endbibitem

\bibitem{akbik-etal-2018-contextual}
\begin{bchapter}
\bauthor{\bsnm{Akbik}, \binits{A.}},
\bauthor{\bsnm{Blythe}, \binits{D.}},
\bauthor{\bsnm{Vollgraf}, \binits{R.}}:
\bctitle{Contextual string embeddings for sequence labeling}.
In: \bbtitle{Proceedings of the 27th International Conference on Computational
  Linguistics},
pp. \bfpage{1638}--\blpage{1649}.
\bpublisher{Association for Computational Linguistics},
\blocation{Santa Fe, New Mexico, USA}
(\byear{2018})
\end{bchapter}
\endbibitem

\bibitem{akbik-etal-2019-pooled}
\begin{bchapter}
\bauthor{\bsnm{Akbik}, \binits{A.}},
\bauthor{\bsnm{Bergmann}, \binits{T.}},
\bauthor{\bsnm{Vollgraf}, \binits{R.}}:
\bctitle{Pooled contextualized embeddings for named entity recognition}.
In: \bbtitle{Proceedings of the 2019 Conference of the North {A}merican Chapter
  of the Association for Computational Linguistics: Human Language
  Technologies, Volume 1 (Long and Short Papers)},
pp. \bfpage{724}--\blpage{728}.
\bpublisher{Association for Computational Linguistics},
\blocation{Minneapolis, Minnesota}
(\byear{2019})
\end{bchapter}
\endbibitem

\bibitem{nguyen2015relation}
\begin{bchapter}
\bauthor{\bsnm{Nguyen}, \binits{T.H.}},
\bauthor{\bsnm{Grishman}, \binits{R.}}:
\bctitle{Relation extraction: Perspective from convolutional neural networks}.
In: \bbtitle{Proceedings of the 1st Workshop on Vector Space Modeling for
  Natural Language Processing},
pp. \bfpage{39}--\blpage{48}
(\byear{2015})
\end{bchapter}
\endbibitem

\bibitem{wick2004}
\begin{barticle}
\bauthor{\bsnm{Wick}, \binits{J.Y.}}:
\batitle{What's in a drug name?: A rose might smell as sweet by any other name,
  but the process of naming the growing number of medications has become quite
  complex and serious.}
\bjtitle{Journal of the American Pharmacists Association}
\bvolume{44}(\bissue{1}),
\bfpage{12}--\blpage{14}
(\byear{2004})
\end{barticle}
\endbibitem

\end{thebibliography}

\newcommand{\BMCxmlcomment}[1]{}

\BMCxmlcomment{

<refgrp>

<bibl id="B1">
  <title><p>Mayo clinical Text Analysis and Knowledge Extraction System
  (cTAKES): architecture, component evaluation and applications</p></title>
  <aug>
    <au><snm>Savova</snm><fnm>GK</fnm></au>
    <au><snm>Masanz</snm><fnm>JJ</fnm></au>
    <au><snm>Ogren</snm><fnm>PV</fnm></au>
    <au><snm>Zheng</snm><fnm>J</fnm></au>
    <au><snm>Sohn</snm><fnm>S</fnm></au>
    <au><snm>Kipper Schuler</snm><fnm>KC</fnm></au>
    <au><snm>Chute</snm><fnm>CG</fnm></au>
  </aug>
  <source>Journal of the American Medical Informatics Association</source>
  <publisher>BMJ Group BMA House, Tavistock Square, London, WC1H
  9JR</publisher>
  <pubdate>2010</pubdate>
  <volume>17</volume>
  <issue>5</issue>
  <fpage>507</fpage>
  <lpage>-513</lpage>
</bibl>

<bibl id="B2">
  <title><p>An overview of MetaMap: historical perspective and recent
  advances</p></title>
  <aug>
    <au><snm>Aronson</snm><fnm>AR</fnm></au>
    <au><snm>Lang</snm><fnm>FM</fnm></au>
  </aug>
  <source>Journal of the American Medical Informatics Association</source>
  <publisher>BMJ Group BMA House, Tavistock Square, London, WC1H
  9JR</publisher>
  <pubdate>2010</pubdate>
  <volume>17</volume>
  <issue>3</issue>
  <fpage>229</fpage>
  <lpage>-236</lpage>
</bibl>

<bibl id="B3">
  <title><p>A general natural-language text processor for clinical
  radiology</p></title>
  <aug>
    <au><snm>Friedman</snm><fnm>C</fnm></au>
    <au><snm>Alderson</snm><fnm>PO</fnm></au>
    <au><snm>Austin</snm><fnm>JH</fnm></au>
    <au><snm>Cimino</snm><fnm>JJ</fnm></au>
    <au><snm>Johnson</snm><fnm>SB</fnm></au>
  </aug>
  <source>Journal of the American Medical Informatics Association</source>
  <publisher>BMJ Group BMA House, Tavistock Square, London, WC1H
  9JR</publisher>
  <pubdate>1994</pubdate>
  <volume>1</volume>
  <issue>2</issue>
  <fpage>161</fpage>
  <lpage>-174</lpage>
</bibl>

<bibl id="B4">
  <title><p>MedEx: a medication information extraction system for clinical
  narratives</p></title>
  <aug>
    <au><snm>Xu</snm><fnm>H</fnm></au>
    <au><snm>Stenner</snm><fnm>SP</fnm></au>
    <au><snm>Doan</snm><fnm>S</fnm></au>
    <au><snm>Johnson</snm><fnm>KB</fnm></au>
    <au><snm>Waitman</snm><fnm>LR</fnm></au>
    <au><snm>Denny</snm><fnm>JC</fnm></au>
  </aug>
  <source>Journal of the American Medical Informatics Association</source>
  <publisher>BMJ Group BMA House, Tavistock Square, London, WC1H
  9JR</publisher>
  <pubdate>2010</pubdate>
  <volume>17</volume>
  <issue>1</issue>
  <fpage>19</fpage>
  <lpage>-24</lpage>
</bibl>

<bibl id="B5">
  <title><p>Comprehensive temporal information detection from clinical text:
  medical events, time, and TLINK identification</p></title>
  <aug>
    <au><snm>Sohn</snm><fnm>S</fnm></au>
    <au><snm>Wagholikar</snm><fnm>KB</fnm></au>
    <au><snm>Li</snm><fnm>D</fnm></au>
    <au><snm>Jonnalagadda</snm><fnm>SR</fnm></au>
    <au><snm>Tao</snm><fnm>C</fnm></au>
    <au><snm>Komandur Elayavilli</snm><fnm>R</fnm></au>
    <au><snm>Liu</snm><fnm>H</fnm></au>
  </aug>
  <source>Journal of the American Medical Informatics Association</source>
  <publisher>BMJ Publishing Group BMA House, Tavistock Square, London, WC1H
  9JR</publisher>
  <pubdate>2013</pubdate>
  <volume>20</volume>
  <issue>5</issue>
  <fpage>836</fpage>
  <lpage>-842</lpage>
</bibl>

<bibl id="B6">
  <title><p>Clinical information extraction applications: A literature
  review</p></title>
  <aug>
    <au><snm>Wang</snm><fnm>Y</fnm></au>
    <au><snm>Wang</snm><fnm>L</fnm></au>
    <au><snm>Rastegar Mojarad</snm><fnm>M</fnm></au>
    <au><snm>Moon</snm><fnm>S</fnm></au>
    <au><snm>Shen</snm><fnm>F</fnm></au>
    <au><snm>Afzal</snm><fnm>N</fnm></au>
    <au><snm>Liu</snm><fnm>S</fnm></au>
    <au><snm>Zeng</snm><fnm>Y</fnm></au>
    <au><snm>Mehrabi</snm><fnm>S</fnm></au>
    <au><snm>Sohn</snm><fnm>S</fnm></au>
    <au><snm>Liu</snm><fnm>H</fnm></au>
  </aug>
  <source>Journal of Biomedical Informatics</source>
  <pubdate>2018</pubdate>
  <volume>77</volume>
  <fpage>34</fpage>
  <lpage>49</lpage>
</bibl>

<bibl id="B7">
  <title><p>Extracting medication information from clinical text</p></title>
  <aug>
    <au><snm>Uzuner</snm><fnm>{\"O}</fnm></au>
    <au><snm>Solti</snm><fnm>I</fnm></au>
    <au><snm>Cadag</snm><fnm>E</fnm></au>
  </aug>
  <source>Journal of the American Medical Informatics Association</source>
  <publisher>Oxford University Press</publisher>
  <pubdate>2010</pubdate>
  <volume>17</volume>
  <issue>5</issue>
  <fpage>514</fpage>
  <lpage>-518</lpage>
</bibl>

<bibl id="B8">
  <title><p>Identifying risk factors for heart disease over time: Overview of
  2014 i2b2/UTHealth shared task Track 2</p></title>
  <aug>
    <au><snm>Stubbs</snm><fnm>A</fnm></au>
    <au><snm>Kotfila</snm><fnm>C</fnm></au>
    <au><snm>Xu</snm><fnm>H</fnm></au>
    <au><snm>Uzuner</snm><fnm>{\"O}</fnm></au>
  </aug>
  <source>Journal of biomedical informatics</source>
  <publisher>Elsevier</publisher>
  <pubdate>2015</pubdate>
  <volume>58</volume>
  <fpage>S67</fpage>
  <lpage>-S77</lpage>
</bibl>

<bibl id="B9">
  <title><p>Overview of the ShARe/CLEF eHealth evaluation lab 2013</p></title>
  <aug>
    <au><snm>Suominen</snm><fnm>H</fnm></au>
    <au><snm>Salanter{\"a}</snm><fnm>S</fnm></au>
    <au><snm>Velupillai</snm><fnm>S</fnm></au>
    <au><snm>Chapman</snm><fnm>WW</fnm></au>
    <au><snm>Savova</snm><fnm>G</fnm></au>
    <au><snm>Elhadad</snm><fnm>N</fnm></au>
    <au><snm>Pradhan</snm><fnm>S</fnm></au>
    <au><snm>South</snm><fnm>BR</fnm></au>
    <au><snm>Mowery</snm><fnm>DL</fnm></au>
    <au><snm>Jones</snm><fnm>GJ</fnm></au>
    <au><cnm>others</cnm></au>
  </aug>
  <source>International Conference of the Cross-Language Evaluation Forum for
  European Languages</source>
  <pubdate>2013</pubdate>
  <fpage>212</fpage>
  <lpage>-231</lpage>
</bibl>

<bibl id="B10">
  <title><p>Overview of the share/clef ehealth evaluation lab 2014</p></title>
  <aug>
    <au><snm>Kelly</snm><fnm>L</fnm></au>
    <au><snm>Goeuriot</snm><fnm>L</fnm></au>
    <au><snm>Suominen</snm><fnm>H</fnm></au>
    <au><snm>Schreck</snm><fnm>T</fnm></au>
    <au><snm>Leroy</snm><fnm>G</fnm></au>
    <au><snm>Mowery</snm><fnm>DL</fnm></au>
    <au><snm>Velupillai</snm><fnm>S</fnm></au>
    <au><snm>Chapman</snm><fnm>WW</fnm></au>
    <au><snm>Martinez</snm><fnm>D</fnm></au>
    <au><snm>Zuccon</snm><fnm>G</fnm></au>
    <au><cnm>others</cnm></au>
  </aug>
  <source>International Conference of the Cross-Language Evaluation Forum for
  European Languages</source>
  <pubdate>2014</pubdate>
  <fpage>172</fpage>
  <lpage>-191</lpage>
</bibl>

<bibl id="B11">
  <title><p>SemEval-2014 Task 7: Analysis of Clinical Text</p></title>
  <aug>
    <au><snm>Pradhan</snm><fnm>S</fnm></au>
    <au><snm>Elhadad</snm><fnm>N</fnm></au>
    <au><snm>Chapman</snm><fnm>W</fnm></au>
    <au><snm>Manandhar</snm><fnm>S</fnm></au>
    <au><snm>Savova</snm><fnm>G</fnm></au>
  </aug>
  <source>Proceedings of the 8th International Workshop on Semantic Evaluation
  (SemEval 2014)</source>
  <pubdate>2014</pubdate>
  <fpage>54</fpage>
  <lpage>-62</lpage>
</bibl>

<bibl id="B12">
  <title><p>Semeval-2016 task 12: Clinical tempeval</p></title>
  <aug>
    <au><snm>Bethard</snm><fnm>S</fnm></au>
    <au><snm>Savova</snm><fnm>G</fnm></au>
    <au><snm>Chen</snm><fnm>WT</fnm></au>
    <au><snm>Derczynski</snm><fnm>L</fnm></au>
    <au><snm>Pustejovsky</snm><fnm>J</fnm></au>
    <au><snm>Verhagen</snm><fnm>M</fnm></au>
  </aug>
  <source>Proceedings of the 10th International Workshop on Semantic Evaluation
  (SemEval-2016)</source>
  <pubdate>2016</pubdate>
  <fpage>1052</fpage>
  <lpage>-1062</lpage>
</bibl>

<bibl id="B13">
  <title><p>CLEF eHealth 2018 Multilingual Information Extraction Task
  Overview: ICD10 Coding of Death Certificates in French, Hungarian and
  Italian.</p></title>
  <aug>
    <au><snm>N{\'e}v{\'e}ol</snm><fnm>A</fnm></au>
    <au><snm>Robert</snm><fnm>A</fnm></au>
    <au><snm>Grippo</snm><fnm>F</fnm></au>
    <au><snm>Morgand</snm><fnm>C</fnm></au>
    <au><snm>Orsi</snm><fnm>C</fnm></au>
    <au><snm>Pelikan</snm><fnm>L</fnm></au>
    <au><snm>Ramadier</snm><fnm>L</fnm></au>
    <au><snm>Rey</snm><fnm>G</fnm></au>
    <au><snm>Zweigenbaum</snm><fnm>P</fnm></au>
  </aug>
  <source>CLEF (Working Notes)</source>
  <pubdate>2018</pubdate>
</bibl>

<bibl id="B14">
  <title><p>Neges 2018: Workshop on negation in spanish</p></title>
  <aug>
    <au><snm>Jim{\'e}nez Zafra</snm><fnm>SM</fnm></au>
    <au><snm>D{\'\i}az</snm><fnm>NPC</fnm></au>
    <au><snm>Morante</snm><fnm>R</fnm></au>
    <au><snm>Mart{\'\i}n Valdivia</snm><fnm>MT</fnm></au>
  </aug>
  <source>Procesamiento del Lenguaje Natural</source>
  <pubdate>2019</pubdate>
  <volume>62</volume>
  <fpage>21</fpage>
  <lpage>-28</lpage>
</bibl>

<bibl id="B15">
  <title><p>Clinical natural language processing in languages other than
  english: opportunities and challenges</p></title>
  <aug>
    <au><snm>N{\'e}v{\'e}ol</snm><fnm>A</fnm></au>
    <au><snm>Dalianis</snm><fnm>H</fnm></au>
    <au><snm>Velupillai</snm><fnm>S</fnm></au>
    <au><snm>Savova</snm><fnm>G</fnm></au>
    <au><snm>Zweigenbaum</snm><fnm>P</fnm></au>
  </aug>
  <source>Journal of biomedical semantics</source>
  <publisher>Springer</publisher>
  <pubdate>2018</pubdate>
  <volume>9</volume>
  <issue>1</issue>
  <fpage>12</fpage>
</bibl>

<bibl id="B16">
  <title><p>Overview of the CLEF eHealth 2019 Multilingual Information
  Extraction</p></title>
  <aug>
    <au><snm>D{\"o}rendahl</snm><fnm>A</fnm></au>
    <au><snm>Leich</snm><fnm>N</fnm></au>
    <au><snm>Hummel</snm><fnm>B</fnm></au>
    <au><snm>Sch{\"o}nfelder</snm><fnm>G</fnm></au>
    <au><snm>Grune</snm><fnm>B</fnm></au>
  </aug>
  <pubdate>2019</pubdate>
</bibl>

<bibl id="B17">
  <title><p>GGPONC: A Corpus of German Medical Text with Rich Metadata Based on
  Clinical Practice Guidelines</p></title>
  <aug>
    <au><snm>Borchert</snm><fnm>F</fnm></au>
    <au><snm>Lohr</snm><fnm>C</fnm></au>
    <au><snm>Modersohn</snm><fnm>L</fnm></au>
    <au><snm>Langer</snm><fnm>T</fnm></au>
    <au><snm>Follmann</snm><fnm>M</fnm></au>
    <au><snm>Sachs</snm><fnm>JP</fnm></au>
    <au><snm>Hahn</snm><fnm>U</fnm></au>
    <au><snm>Schapranow</snm><fnm>MP</fnm></au>
  </aug>
  <source>Proceedings of the 11th International Workshop on Health Text Mining
  and Information Analysis</source>
  <pubdate>2020</pubdate>
  <fpage>38</fpage>
  <lpage>-48</lpage>
</bibl>

<bibl id="B18">
  <title><p>From Witch’s Shot to Music Making Bones-Resources for Medical
  Laymen to Technical Language and Vice Versa</p></title>
  <aug>
    <au><snm>Seiffe</snm><fnm>L</fnm></au>
    <au><snm>Marten</snm><fnm>O</fnm></au>
    <au><snm>Mikhailov</snm><fnm>M</fnm></au>
    <au><snm>Schmeier</snm><fnm>S</fnm></au>
    <au><snm>M{\"o}ller</snm><fnm>S</fnm></au>
    <au><snm>Roller</snm><fnm>R</fnm></au>
  </aug>
  <source>Proceedings of the 12th Language Resources and Evaluation
  Conference</source>
  <pubdate>2020</pubdate>
  <fpage>6185</fpage>
  <lpage>-6192</lpage>
</bibl>

<bibl id="B19">
  <title><p>Sharing Copies of Synthetic Clinical Corpora without Physical
  Distribution—A Case Study to Get Around IPRs and Privacy Constraints
  Featuring the German JSYNCC Corpus</p></title>
  <aug>
    <au><snm>Lohr</snm><fnm>C</fnm></au>
    <au><snm>Buechel</snm><fnm>S</fnm></au>
    <au><snm>Hahn</snm><fnm>U</fnm></au>
  </aug>
  <source>Proceedings of the Eleventh International Conference on Language
  Resources and Evaluation (LREC 2018)</source>
  <pubdate>2018</pubdate>
</bibl>

<bibl id="B20">
  <title><p>{Sharing Models and Tools for Processing German Clinical
  Texts}</p></title>
  <aug>
    <au><snm>Hellrich</snm><fnm>J</fnm></au>
    <au><snm>Matthies</snm><fnm>F</fnm></au>
    <au><snm>Faessler</snm><fnm>E</fnm></au>
    <au><snm>Hahn</snm><fnm>U</fnm></au>
  </aug>
  <source>MIE 2015 - Digital Healthcare Empowering Europeans</source>
  <publisher>Studies in Health Technology and Informatics</publisher>
  <pubdate>2015</pubdate>
  <volume>210</volume>
  <fpage>734</fpage>
  <lpage>-738</lpage>
</bibl>

<bibl id="B21">
  <title><p>Really, is medical sublanguage that different? Experimental
  counter-evidence from tagging medical and newspaper corpora.</p></title>
  <aug>
    <au><snm>Wermter</snm><fnm>J</fnm></au>
    <au><snm>Hahn</snm><fnm>U</fnm></au>
  </aug>
  <source>Medinfo</source>
  <pubdate>2004</pubdate>
  <fpage>560</fpage>
  <lpage>-564</lpage>
</bibl>

<bibl id="B22">
  <title><p>A Simple Algorithm for Identifying Negated Findings and Diseases in
  Discharge Summaries</p></title>
  <aug>
    <au><snm>Chapman</snm><fnm>WW</fnm></au>
    <au><snm>Bridewell</snm><fnm>W</fnm></au>
    <au><snm>Hanbury</snm><fnm>P</fnm></au>
    <au><snm>Cooper</snm><fnm>GF</fnm></au>
    <au><snm>Buchanan</snm><fnm>BG</fnm></au>
  </aug>
  <source>Journal of Biomedical Informatics</source>
  <pubdate>2001</pubdate>
  <volume>34</volume>
  <issue>5</issue>
  <fpage>301</fpage>
  <lpage>310</lpage>
</bibl>

<bibl id="B23">
  <title><p>Extending the NegEx lexicon for multiple languages</p></title>
  <aug>
    <au><snm>Chapman</snm><fnm>WW</fnm></au>
    <au><snm>Hilert</snm><fnm>D</fnm></au>
    <au><snm>Velupillai</snm><fnm>S</fnm></au>
    <au><snm>Kvist</snm><fnm>M</fnm></au>
    <au><snm>Skeppstedt</snm><fnm>M</fnm></au>
    <au><snm>Chapman</snm><fnm>BE</fnm></au>
    <au><snm>Conway</snm><fnm>M</fnm></au>
    <au><snm>Tharp</snm><fnm>M</fnm></au>
    <au><snm>Mowery</snm><fnm>DL</fnm></au>
    <au><snm>Deleger</snm><fnm>L</fnm></au>
  </aug>
  <source>Studies in health technology and informatics</source>
  <publisher>NIH Public Access</publisher>
  <pubdate>2013</pubdate>
  <volume>192</volume>
  <fpage>677</fpage>
</bibl>

<bibl id="B24">
  <title><p>Negation Detection in Clinical Reports Written in
  {G}erman</p></title>
  <aug>
    <au><snm>Cotik</snm><fnm>V</fnm></au>
    <au><snm>Roller</snm><fnm>R</fnm></au>
    <au><snm>Xu</snm><fnm>F</fnm></au>
    <au><snm>Uszkoreit</snm><fnm>H</fnm></au>
    <au><snm>Budde</snm><fnm>K</fnm></au>
    <au><snm>Schmidt</snm><fnm>D</fnm></au>
  </aug>
  <source>Proceedings of the Fifth Workshop on Building and Evaluating
  Resources for Biomedical Text Mining ({B}io{T}xt{M}2016)</source>
  <publisher>Osaka, Japan: The COLING 2016 Organizing Committee</publisher>
  <pubdate>2016</pubdate>
  <fpage>115</fpage>
  <lpage>-124</lpage>
</bibl>

<bibl id="B25">
  <title><p>{A Domain-adapted Dependency Parser for German Clinical
  Text}</p></title>
  <aug>
    <au><snm>Kara</snm><fnm>E</fnm></au>
    <au><snm>Zeen</snm><fnm>T</fnm></au>
    <au><snm>Gabryszak</snm><fnm>A</fnm></au>
    <au><snm>Budde</snm><fnm>K</fnm></au>
    <au><snm>Schmidt</snm><fnm>D</fnm></au>
    <au><snm>Roller</snm><fnm>R</fnm></au>
  </aug>
  <source>Proceedings of the 14th Conference on Natural Language Processing
  (KONVENS 2018)</source>
  <publisher>Vienna, Austria</publisher>
  <pubdate>2018</pubdate>
</bibl>

<bibl id="B26">
  <title><p>Unsupervised Abbreviation Expansion in Clinical
  Narratives.</p></title>
  <aug>
    <au><snm>Oleynik</snm><fnm>M</fnm></au>
    <au><snm>Kreuzthaler</snm><fnm>M</fnm></au>
    <au><snm>Schulz</snm><fnm>S</fnm></au>
  </aug>
  <source>MedInfo</source>
  <pubdate>2017</pubdate>
  <volume>245</volume>
  <fpage>539</fpage>
  <lpage>-543</lpage>
</bibl>

<bibl id="B27">
  <title><p>{Pseudonymization of PHI items in German clinical
  reports}</p></title>
  <aug>
    <au><snm>Lohr</snm><fnm>C</fnm></au>
    <au><snm>Eder</snm><fnm>E</fnm></au>
    <au><snm>Hahn</snm><fnm>U</fnm></au>
  </aug>
  <source>Public Health and Informatics</source>
  <pubdate>2021</pubdate>
</bibl>

<bibl id="B28">
  <title><p>Annotation and initial evaluation of a large annotated German
  oncological corpus</p></title>
  <aug>
    <au><snm>Kittner</snm><fnm>M</fnm></au>
    <au><snm>Lamping</snm><fnm>M</fnm></au>
    <au><snm>Rieke</snm><fnm>DT</fnm></au>
    <au><snm>G{\"o}tze</snm><fnm>J</fnm></au>
    <au><snm>Bajwa</snm><fnm>B</fnm></au>
    <au><snm>Jelas</snm><fnm>I</fnm></au>
    <au><snm>R{\"u}ter</snm><fnm>G</fnm></au>
    <au><snm>Hautow</snm><fnm>H</fnm></au>
    <au><snm>S{\"a}nger</snm><fnm>M</fnm></au>
    <au><snm>Habibi</snm><fnm>M</fnm></au>
    <au><cnm>others</cnm></au>
  </aug>
  <source>JAMIA open</source>
  <publisher>Oxford University Press</publisher>
  <pubdate>2021</pubdate>
  <volume>4</volume>
  <issue>2</issue>
  <fpage>ooab025</fpage>
</bibl>

<bibl id="B29">
  <title><p>GERNERMED: An open German medical NER model</p></title>
  <aug>
    <au><snm>Frei</snm><fnm>J</fnm></au>
    <au><snm>Kramer</snm><fnm>F</fnm></au>
  </aug>
  <source>Software Impacts</source>
  <publisher>Elsevier</publisher>
  <pubdate>2022</pubdate>
  <volume>11</volume>
  <fpage>100212</fpage>
</bibl>

<bibl id="B30">
  <title><p>2018 n2c2 shared task on adverse drug events and medication
  extraction in electronic health records</p></title>
  <aug>
    <au><snm>Henry</snm><fnm>S</fnm></au>
    <au><snm>Buchan</snm><fnm>K</fnm></au>
    <au><snm>Filannino</snm><fnm>M</fnm></au>
    <au><snm>Stubbs</snm><fnm>A</fnm></au>
    <au><snm>Uzuner</snm><fnm>O</fnm></au>
  </aug>
  <source>Journal of the American Medical Informatics Association</source>
  <publisher>Oxford University Press</publisher>
  <pubdate>2020</pubdate>
  <volume>27</volume>
  <issue>1</issue>
  <fpage>3</fpage>
  <lpage>-12</lpage>
</bibl>

<bibl id="B31">
  <title><p>Development of a language model for medical domain</p></title>
  <aug>
    <au><snm>Shrestha</snm><fnm>M</fnm></au>
  </aug>
  <source>PhD thesis</source>
  <publisher>Hochschule Rhein-Waal</publisher>
  <pubdate>2021</pubdate>
</bibl>

<bibl id="B32">
  <title><p>German-MedBERT on Hugging Face</p></title>
  <source>\url{https://huggingface.co/smanjil/German-MedBERT}</source>
  <note>Accessed: 2022-03-15</note>
</bibl>

<bibl id="B33">
  <title><p>Semi-automated De-identification of German Content Sensitive
  Reports for Big Data Analytics</p></title>
  <aug>
    <au><snm>Seuss</snm><fnm>H</fnm></au>
    <au><snm>Dankerl</snm><fnm>P</fnm></au>
    <au><snm>Ihle</snm><fnm>M</fnm></au>
    <au><snm>Grandjean</snm><fnm>A</fnm></au>
    <au><snm>Hammon</snm><fnm>R</fnm></au>
    <au><snm>Kaestle</snm><fnm>N</fnm></au>
    <au><snm>Fasching</snm><fnm>PA</fnm></au>
    <au><snm>Maier</snm><fnm>C</fnm></au>
    <au><snm>Christoph</snm><fnm>J</fnm></au>
    <au><snm>Sedlmayr</snm><fnm>M</fnm></au>
    <au><cnm>others</cnm></au>
  </aug>
  <source>R{\"o}Fo-Fortschritte auf dem Gebiet der R{\"o}ntgenstrahlen und der
  bildgebenden Verfahren</source>
  <pubdate>2017</pubdate>
  <volume>189</volume>
  <issue>07</issue>
  <fpage>661</fpage>
  <lpage>-671</lpage>
</bibl>

<bibl id="B34">
  <title><p>Beckers Abkürzungslexikon Medizinischer Begriffe</p></title>
  <source>\url{https://www.medizinische-abkuerzungen.de/suche.html}</source>
  <note>Accessed: 2021-01-10</note>
</bibl>

<bibl id="B35">
  <title><p>A fine-grained corpus annotation schema of German nephrology
  records</p></title>
  <aug>
    <au><snm>Roller</snm><fnm>R</fnm></au>
    <au><snm>Uszkoreit</snm><fnm>H</fnm></au>
    <au><snm>Xu</snm><fnm>F</fnm></au>
    <au><snm>Seiffe</snm><fnm>L</fnm></au>
    <au><snm>Mikhailov</snm><fnm>M</fnm></au>
    <au><snm>Staeck</snm><fnm>O</fnm></au>
    <au><snm>Budde</snm><fnm>K</fnm></au>
    <au><snm>Halleck</snm><fnm>F</fnm></au>
    <au><snm>Schmidt</snm><fnm>D</fnm></au>
  </aug>
  <source>Proceedings of the Clinical Natural Language Processing Workshop
  (ClinicalNLP)</source>
  <publisher>Osaka, Japan: The COLING 2016 Organizing Committee</publisher>
  <pubdate>2016</pubdate>
  <fpage>69</fpage>
  <lpage>-77</lpage>
</bibl>

<bibl id="B36">
  <title><p>{brat}: a Web-based Tool for {NLP}-Assisted Text
  Annotation</p></title>
  <aug>
    <au><snm>Stenetorp</snm><fnm>P</fnm></au>
    <au><snm>Pyysalo</snm><fnm>S</fnm></au>
    <au><snm>Topi\'{c}</snm><fnm>G</fnm></au>
    <au><snm>Ohta</snm><fnm>T</fnm></au>
    <au><snm>Ananiadou</snm><fnm>S</fnm></au>
    <au><snm>Tsujii</snm><fnm>J</fnm></au>
  </aug>
  <source>Proceedings of the Demonstrations Session at {EACL} 2012</source>
  <publisher>Avignon, France: Association for Computational
  Linguistics</publisher>
  <pubdate>2012</pubdate>
</bibl>

<bibl id="B37">
  <title><p>Agreement, the f-measure, and reliability in information
  retrieval</p></title>
  <aug>
    <au><snm>Hripcsak</snm><fnm>G</fnm></au>
    <au><snm>Rothschild</snm><fnm>AS</fnm></au>
  </aug>
  <source>Journal of the American medical informatics association</source>
  <publisher>BMJ Group BMA House, Tavistock Square, London, WC1H
  9JR</publisher>
  <pubdate>2005</pubdate>
  <volume>12</volume>
  <issue>3</issue>
  <fpage>296</fpage>
  <lpage>-298</lpage>
</bibl>

<bibl id="B38">
  <title><p>{Linguistic Modeling for Text Analytic Tasks for German Clinical
  Texts}</p></title>
  <aug>
    <au><snm>Seiffe</snm><fnm>L</fnm></au>
  </aug>
  <source>Master's thesis</source>
  <publisher>TU Berlin</publisher>
  <pubdate>2018</pubdate>
</bibl>

<bibl id="B39">
  <title><p>{Guidelines f{\"u}r das Tagging deutscher Textcorpora mit
  STTS}</p></title>
  <aug>
    <au><snm>Schiller</snm><fnm>A</fnm></au>
    <au><snm>Teufel</snm><fnm>S</fnm></au>
    <au><snm>Thielen</snm><fnm>C</fnm></au>
  </aug>
  <source>Universit{\"a}ten Stuttgart und T{\"u}bingen</source>
  <pubdate>1999</pubdate>
</bibl>

<bibl id="B40">
  <title><p>Universal dependencies v1: A multilingual treebank
  collection</p></title>
  <aug>
    <au><snm>Nivre</snm><fnm>J</fnm></au>
    <au><snm>De Marneffe</snm><fnm>MC</fnm></au>
    <au><snm>Ginter</snm><fnm>F</fnm></au>
    <au><snm>Goldberg</snm><fnm>Y</fnm></au>
    <au><snm>Hajic</snm><fnm>J</fnm></au>
    <au><snm>Manning</snm><fnm>CD</fnm></au>
    <au><snm>McDonald</snm><fnm>R</fnm></au>
    <au><snm>Petrov</snm><fnm>S</fnm></au>
    <au><snm>Pyysalo</snm><fnm>S</fnm></au>
    <au><snm>Silveira</snm><fnm>N</fnm></au>
    <au><cnm>others</cnm></au>
  </aug>
  <source>Proceedings of the Tenth International Conference on Language
  Resources and Evaluation (LREC'16)</source>
  <pubdate>2016</pubdate>
  <fpage>1659</fpage>
  <lpage>-1666</lpage>
</bibl>

<bibl id="B41">
  <title><p>Because Size Does Matter: The Hamburg Dependency
  Treebank</p></title>
  <aug>
    <au><snm>Foth</snm><fnm>KA</fnm></au>
    <au><snm>Köhn</snm><fnm>A</fnm></au>
    <au><snm>Beuck</snm><fnm>N</fnm></au>
    <au><snm>Menzel</snm><fnm>W</fnm></au>
  </aug>
  <source>Proceedings of the Ninth International Conference on Language
  Resources and Evaluation (LREC'14)</source>
  <publisher>Reykjavik, Iceland: European Language Resources Association
  (ELRA)</publisher>
  <editor>Nicoletta Calzolari (Conference Chair) and Khalid Choukri and Thierry
  Declerck and Hrafn Loftsson and Bente Maegaard and Joseph Mariani and
  Asuncion Moreno and Jan Odijk and Stelios Piperidis</editor>
  <pubdate>2014</pubdate>
</bibl>

<bibl id="B42">
  <title><p>Med1 Forum</p></title>
  <source>\url{https://med2-forum.de/}</source>
  <note>Accessed: 2021-01-10; Note, the Med1 forum does not exist anymore, see
  med2 instead.</note>
</bibl>

<bibl id="B43">
  <title><p>Deutsches Medizin Forum</p></title>
  <source>\url{https://www.medizin-forum.de/}</source>
  <note>Accessed: 2021-01-10</note>
</bibl>

<bibl id="B44">
  <title><p>Spiegel</p></title>
  <source>\url{https://www.spiegel.de/gesundheit/}</source>
  <note>Accessed: 2021-01-10</note>
</bibl>

<bibl id="B45">
  <title><p>Aerzte-Blatt</p></title>
  <source>\url{https://www.aerzteblatt.de/}</source>
  <note>Accessed: 2021-01-10</note>
</bibl>

<bibl id="B46">
  <title><p>NetDoktor</p></title>
  <source>\url{https://www.netdoktor.de/}</source>
  <note>Accessed: 2021-01-10</note>
</bibl>

<bibl id="B47">
  <title><p>Onmeda</p></title>
  <source>\url{https://www.onmeda.de/}</source>
  <note>Accessed: 2021-01-10</note>
</bibl>

<bibl id="B48">
  <title><p>German PubMed Abstracts</p></title>
  <source>\url{https://pubmed.ncbi.nlm.nih.gov/}</source>
  <note>Accessed: 2021-01-10</note>
</bibl>

<bibl id="B49">
  <title><p>eDocTrainer</p></title>
  <source>\url{https://www.edoctrainer.de/}</source>
  <note>Accessed: 2021-01-10</note>
</bibl>

<bibl id="B50">
  <title><p>Chirurgie: mit integriertem Fallquiz - 40 F{\"a}lle nach neuer
  AO</p></title>
  <aug>
    <au><snm>Siewert</snm><fnm>J.R.</fnm></au>
    <au><snm>Stein</snm><fnm>H.J.</fnm></au>
    <au><snm>Allg{\"o}wer</snm><fnm>M.</fnm></au>
  </aug>
  <publisher>Springer Verlag Berlin Heidelberg</publisher>
  <pubdate>2009</pubdate>
</bibl>

<bibl id="B51">
  <title><p>Neurologie</p></title>
  <aug>
    <au><snm>Poeck</snm><fnm>K.</fnm></au>
    <au><snm>Hacke</snm><fnm>W.</fnm></au>
  </aug>
  <publisher>Springer Verlag Berlin Heidelberg</publisher>
  <pubdate>2006</pubdate>
</bibl>

<bibl id="B52">
  <title><p>Urologie</p></title>
  <aug>
    <au><snm>Hautmann</snm><fnm>R.</fnm></au>
    <au><snm>Gschwend</snm><fnm>J.E.</fnm></au>
  </aug>
  <publisher>Springer Verlag Berlin Heidelberg</publisher>
  <pubdate>2014</pubdate>
</bibl>

<bibl id="B53">
  <title><p>Basiswissen Augenheilkunde</p></title>
  <aug>
    <au><snm>Walter</snm><fnm>P</fnm></au>
    <au><snm>Plange</snm><fnm>N</fnm></au>
  </aug>
  <publisher>Springer</publisher>
  <pubdate>2017</pubdate>
</bibl>

<bibl id="B54">
  <title><p>Hals-Nasen-Ohren-Heilkunde</p></title>
  <aug>
    <au><snm>Lenarz</snm><fnm>T.</fnm></au>
    <au><snm>Boenninghaus</snm><fnm>H.G.</fnm></au>
  </aug>
  <publisher>Springer Verlag Berlin Heidelberg</publisher>
  <pubdate>2012</pubdate>
</bibl>

<bibl id="B55">
  <title><p>Notfallmedizin</p></title>
  <aug>
    <au><snm>Ziegenfu{\ss}</snm><fnm>T.</fnm></au>
  </aug>
  <publisher>Springer Verlag Berlin Heidelberg</publisher>
  <pubdate>2016</pubdate>
</bibl>

<bibl id="B56">
  <title><p>Basiswissen Dermatologie</p></title>
  <aug>
    <au><snm>Goebeler</snm><fnm>M.</fnm></au>
    <au><snm>Hamm</snm><fnm>H.</fnm></au>
  </aug>
  <publisher>Springer Verlag Berlin Heidelberg</publisher>
  <pubdate>2017</pubdate>
</bibl>

<bibl id="B57">
  <title><p>Basiswissen Psychiatrie und Psychotherapie</p></title>
  <aug>
    <au><snm>Arolt</snm><fnm>V</fnm></au>
    <au><snm>Reimer</snm><fnm>C</fnm></au>
    <au><snm>Dilling</snm><fnm>H</fnm></au>
  </aug>
  <publisher>Springer-Verlag</publisher>
  <pubdate>2011</pubdate>
</bibl>

<bibl id="B58">
  <title><p>Enriching word vectors with subword information</p></title>
  <aug>
    <au><snm>Bojanowski</snm><fnm>P</fnm></au>
    <au><snm>Grave</snm><fnm>E</fnm></au>
    <au><snm>Joulin</snm><fnm>A</fnm></au>
    <au><snm>Mikolov</snm><fnm>T</fnm></au>
  </aug>
  <source>Transactions of the association for computational
  linguistics</source>
  <publisher>MIT Press</publisher>
  <pubdate>2017</pubdate>
  <volume>5</volume>
  <fpage>135</fpage>
  <lpage>-146</lpage>
</bibl>

<bibl id="B59">
  <title><p>Contextual String Embeddings for Sequence Labeling</p></title>
  <aug>
    <au><snm>Akbik</snm><fnm>A</fnm></au>
    <au><snm>Blythe</snm><fnm>D</fnm></au>
    <au><snm>Vollgraf</snm><fnm>R</fnm></au>
  </aug>
  <source>Proceedings of the 27th International Conference on Computational
  Linguistics</source>
  <publisher>Santa Fe, New Mexico, USA: Association for Computational
  Linguistics</publisher>
  <pubdate>2018</pubdate>
  <fpage>1638</fpage>
  <lpage>-1649</lpage>
</bibl>

<bibl id="B60">
  <title><p>Pooled Contextualized Embeddings for Named Entity
  Recognition</p></title>
  <aug>
    <au><snm>Akbik</snm><fnm>A</fnm></au>
    <au><snm>Bergmann</snm><fnm>T</fnm></au>
    <au><snm>Vollgraf</snm><fnm>R</fnm></au>
  </aug>
  <source>Proceedings of the 2019 Conference of the North {A}merican Chapter of
  the Association for Computational Linguistics: Human Language Technologies,
  Volume 1 (Long and Short Papers)</source>
  <publisher>Minneapolis, Minnesota: Association for Computational
  Linguistics</publisher>
  <pubdate>2019</pubdate>
  <fpage>724</fpage>
  <lpage>-728</lpage>
</bibl>

<bibl id="B61">
  <title><p>Relation extraction: Perspective from convolutional neural
  networks</p></title>
  <aug>
    <au><snm>Nguyen</snm><fnm>TH</fnm></au>
    <au><snm>Grishman</snm><fnm>R</fnm></au>
  </aug>
  <source>Proceedings of the 1st Workshop on Vector Space Modeling for Natural
  Language Processing</source>
  <pubdate>2015</pubdate>
  <fpage>39</fpage>
  <lpage>-48</lpage>
</bibl>

<bibl id="B62">
  <title><p>What's in a Drug Name?: A rose might smell as sweet by any other
  name, but the process of naming the growing number of medications has become
  quite complex and serious.</p></title>
  <aug>
    <au><snm>Wick</snm><fnm>JY</fnm></au>
  </aug>
  <source>Journal of the American Pharmacists Association</source>
  <pubdate>2004</pubdate>
  <volume>44</volume>
  <issue>1</issue>
  <fpage>12</fpage>
  <lpage>-14</lpage>
</bibl>

</refgrp>
} 





\newcommand{\BMCxmlcomment}[1]{}

\BMCxmlcomment{

<refgrp>

<bibl id="B1">
  <title><p>Mayo clinical Text Analysis and Knowledge Extraction System
  (cTAKES): architecture, component evaluation and applications</p></title>
  <aug>
    <au><snm>Savova</snm><fnm>GK</fnm></au>
    <au><snm>Masanz</snm><fnm>JJ</fnm></au>
    <au><snm>Ogren</snm><fnm>PV</fnm></au>
    <au><snm>Zheng</snm><fnm>J</fnm></au>
    <au><snm>Sohn</snm><fnm>S</fnm></au>
    <au><snm>Kipper Schuler</snm><fnm>KC</fnm></au>
    <au><snm>Chute</snm><fnm>CG</fnm></au>
  </aug>
  <source>Journal of the American Medical Informatics Association</source>
  <publisher>BMJ Group BMA House, Tavistock Square, London, WC1H
  9JR</publisher>
  <pubdate>2010</pubdate>
  <volume>17</volume>
  <issue>5</issue>
  <fpage>507</fpage>
  <lpage>-513</lpage>
</bibl>

<bibl id="B2">
  <title><p>An overview of MetaMap: historical perspective and recent
  advances</p></title>
  <aug>
    <au><snm>Aronson</snm><fnm>AR</fnm></au>
    <au><snm>Lang</snm><fnm>FM</fnm></au>
  </aug>
  <source>Journal of the American Medical Informatics Association</source>
  <publisher>BMJ Group BMA House, Tavistock Square, London, WC1H
  9JR</publisher>
  <pubdate>2010</pubdate>
  <volume>17</volume>
  <issue>3</issue>
  <fpage>229</fpage>
  <lpage>-236</lpage>
</bibl>

<bibl id="B3">
  <title><p>A general natural-language text processor for clinical
  radiology</p></title>
  <aug>
    <au><snm>Friedman</snm><fnm>C</fnm></au>
    <au><snm>Alderson</snm><fnm>PO</fnm></au>
    <au><snm>Austin</snm><fnm>JH</fnm></au>
    <au><snm>Cimino</snm><fnm>JJ</fnm></au>
    <au><snm>Johnson</snm><fnm>SB</fnm></au>
  </aug>
  <source>Journal of the American Medical Informatics Association</source>
  <publisher>BMJ Group BMA House, Tavistock Square, London, WC1H
  9JR</publisher>
  <pubdate>1994</pubdate>
  <volume>1</volume>
  <issue>2</issue>
  <fpage>161</fpage>
  <lpage>-174</lpage>
</bibl>

<bibl id="B4">
  <title><p>MedEx: a medication information extraction system for clinical
  narratives</p></title>
  <aug>
    <au><snm>Xu</snm><fnm>H</fnm></au>
    <au><snm>Stenner</snm><fnm>SP</fnm></au>
    <au><snm>Doan</snm><fnm>S</fnm></au>
    <au><snm>Johnson</snm><fnm>KB</fnm></au>
    <au><snm>Waitman</snm><fnm>LR</fnm></au>
    <au><snm>Denny</snm><fnm>JC</fnm></au>
  </aug>
  <source>Journal of the American Medical Informatics Association</source>
  <publisher>BMJ Group BMA House, Tavistock Square, London, WC1H
  9JR</publisher>
  <pubdate>2010</pubdate>
  <volume>17</volume>
  <issue>1</issue>
  <fpage>19</fpage>
  <lpage>-24</lpage>
</bibl>

<bibl id="B5">
  <title><p>Comprehensive temporal information detection from clinical text:
  medical events, time, and TLINK identification</p></title>
  <aug>
    <au><snm>Sohn</snm><fnm>S</fnm></au>
    <au><snm>Wagholikar</snm><fnm>KB</fnm></au>
    <au><snm>Li</snm><fnm>D</fnm></au>
    <au><snm>Jonnalagadda</snm><fnm>SR</fnm></au>
    <au><snm>Tao</snm><fnm>C</fnm></au>
    <au><snm>Komandur Elayavilli</snm><fnm>R</fnm></au>
    <au><snm>Liu</snm><fnm>H</fnm></au>
  </aug>
  <source>Journal of the American Medical Informatics Association</source>
  <publisher>BMJ Publishing Group BMA House, Tavistock Square, London, WC1H
  9JR</publisher>
  <pubdate>2013</pubdate>
  <volume>20</volume>
  <issue>5</issue>
  <fpage>836</fpage>
  <lpage>-842</lpage>
</bibl>

<bibl id="B6">
  <title><p>Clinical information extraction applications: A literature
  review</p></title>
  <aug>
    <au><snm>Wang</snm><fnm>Y</fnm></au>
    <au><snm>Wang</snm><fnm>L</fnm></au>
    <au><snm>Rastegar Mojarad</snm><fnm>M</fnm></au>
    <au><snm>Moon</snm><fnm>S</fnm></au>
    <au><snm>Shen</snm><fnm>F</fnm></au>
    <au><snm>Afzal</snm><fnm>N</fnm></au>
    <au><snm>Liu</snm><fnm>S</fnm></au>
    <au><snm>Zeng</snm><fnm>Y</fnm></au>
    <au><snm>Mehrabi</snm><fnm>S</fnm></au>
    <au><snm>Sohn</snm><fnm>S</fnm></au>
    <au><snm>Liu</snm><fnm>H</fnm></au>
  </aug>
  <source>Journal of Biomedical Informatics</source>
  <pubdate>2018</pubdate>
  <volume>77</volume>
  <fpage>34</fpage>
  <lpage>49</lpage>
</bibl>

<bibl id="B7">
  <title><p>Extracting medication information from clinical text</p></title>
  <aug>
    <au><snm>Uzuner</snm><fnm>{\"O}</fnm></au>
    <au><snm>Solti</snm><fnm>I</fnm></au>
    <au><snm>Cadag</snm><fnm>E</fnm></au>
  </aug>
  <source>Journal of the American Medical Informatics Association</source>
  <publisher>Oxford University Press</publisher>
  <pubdate>2010</pubdate>
  <volume>17</volume>
  <issue>5</issue>
  <fpage>514</fpage>
  <lpage>-518</lpage>
</bibl>

<bibl id="B8">
  <title><p>Identifying risk factors for heart disease over time: Overview of
  2014 i2b2/UTHealth shared task Track 2</p></title>
  <aug>
    <au><snm>Stubbs</snm><fnm>A</fnm></au>
    <au><snm>Kotfila</snm><fnm>C</fnm></au>
    <au><snm>Xu</snm><fnm>H</fnm></au>
    <au><snm>Uzuner</snm><fnm>{\"O}</fnm></au>
  </aug>
  <source>Journal of biomedical informatics</source>
  <publisher>Elsevier</publisher>
  <pubdate>2015</pubdate>
  <volume>58</volume>
  <fpage>S67</fpage>
  <lpage>-S77</lpage>
</bibl>

<bibl id="B9">
  <title><p>Overview of the ShARe/CLEF eHealth evaluation lab 2013</p></title>
  <aug>
    <au><snm>Suominen</snm><fnm>H</fnm></au>
    <au><snm>Salanter{\"a}</snm><fnm>S</fnm></au>
    <au><snm>Velupillai</snm><fnm>S</fnm></au>
    <au><snm>Chapman</snm><fnm>WW</fnm></au>
    <au><snm>Savova</snm><fnm>G</fnm></au>
    <au><snm>Elhadad</snm><fnm>N</fnm></au>
    <au><snm>Pradhan</snm><fnm>S</fnm></au>
    <au><snm>South</snm><fnm>BR</fnm></au>
    <au><snm>Mowery</snm><fnm>DL</fnm></au>
    <au><snm>Jones</snm><fnm>GJ</fnm></au>
    <au><cnm>others</cnm></au>
  </aug>
  <source>International Conference of the Cross-Language Evaluation Forum for
  European Languages</source>
  <pubdate>2013</pubdate>
  <fpage>212</fpage>
  <lpage>-231</lpage>
</bibl>

<bibl id="B10">
  <title><p>Overview of the share/clef ehealth evaluation lab 2014</p></title>
  <aug>
    <au><snm>Kelly</snm><fnm>L</fnm></au>
    <au><snm>Goeuriot</snm><fnm>L</fnm></au>
    <au><snm>Suominen</snm><fnm>H</fnm></au>
    <au><snm>Schreck</snm><fnm>T</fnm></au>
    <au><snm>Leroy</snm><fnm>G</fnm></au>
    <au><snm>Mowery</snm><fnm>DL</fnm></au>
    <au><snm>Velupillai</snm><fnm>S</fnm></au>
    <au><snm>Chapman</snm><fnm>WW</fnm></au>
    <au><snm>Martinez</snm><fnm>D</fnm></au>
    <au><snm>Zuccon</snm><fnm>G</fnm></au>
    <au><cnm>others</cnm></au>
  </aug>
  <source>International Conference of the Cross-Language Evaluation Forum for
  European Languages</source>
  <pubdate>2014</pubdate>
  <fpage>172</fpage>
  <lpage>-191</lpage>
</bibl>

<bibl id="B11">
  <title><p>SemEval-2014 Task 7: Analysis of Clinical Text</p></title>
  <aug>
    <au><snm>Pradhan</snm><fnm>S</fnm></au>
    <au><snm>Elhadad</snm><fnm>N</fnm></au>
    <au><snm>Chapman</snm><fnm>W</fnm></au>
    <au><snm>Manandhar</snm><fnm>S</fnm></au>
    <au><snm>Savova</snm><fnm>G</fnm></au>
  </aug>
  <source>Proceedings of the 8th International Workshop on Semantic Evaluation
  (SemEval 2014)</source>
  <pubdate>2014</pubdate>
  <fpage>54</fpage>
  <lpage>-62</lpage>
</bibl>

<bibl id="B12">
  <title><p>Semeval-2016 task 12: Clinical tempeval</p></title>
  <aug>
    <au><snm>Bethard</snm><fnm>S</fnm></au>
    <au><snm>Savova</snm><fnm>G</fnm></au>
    <au><snm>Chen</snm><fnm>WT</fnm></au>
    <au><snm>Derczynski</snm><fnm>L</fnm></au>
    <au><snm>Pustejovsky</snm><fnm>J</fnm></au>
    <au><snm>Verhagen</snm><fnm>M</fnm></au>
  </aug>
  <source>Proceedings of the 10th International Workshop on Semantic Evaluation
  (SemEval-2016)</source>
  <pubdate>2016</pubdate>
  <fpage>1052</fpage>
  <lpage>-1062</lpage>
</bibl>

<bibl id="B13">
  <title><p>CLEF eHealth 2018 Multilingual Information Extraction Task
  Overview: ICD10 Coding of Death Certificates in French, Hungarian and
  Italian.</p></title>
  <aug>
    <au><snm>N{\'e}v{\'e}ol</snm><fnm>A</fnm></au>
    <au><snm>Robert</snm><fnm>A</fnm></au>
    <au><snm>Grippo</snm><fnm>F</fnm></au>
    <au><snm>Morgand</snm><fnm>C</fnm></au>
    <au><snm>Orsi</snm><fnm>C</fnm></au>
    <au><snm>Pelikan</snm><fnm>L</fnm></au>
    <au><snm>Ramadier</snm><fnm>L</fnm></au>
    <au><snm>Rey</snm><fnm>G</fnm></au>
    <au><snm>Zweigenbaum</snm><fnm>P</fnm></au>
  </aug>
  <source>CLEF (Working Notes)</source>
  <pubdate>2018</pubdate>
</bibl>

<bibl id="B14">
  <title><p>Neges 2018: Workshop on negation in spanish</p></title>
  <aug>
    <au><snm>Jim{\'e}nez Zafra</snm><fnm>SM</fnm></au>
    <au><snm>D{\'\i}az</snm><fnm>NPC</fnm></au>
    <au><snm>Morante</snm><fnm>R</fnm></au>
    <au><snm>Mart{\'\i}n Valdivia</snm><fnm>MT</fnm></au>
  </aug>
  <source>Procesamiento del Lenguaje Natural</source>
  <pubdate>2019</pubdate>
  <volume>62</volume>
  <fpage>21</fpage>
  <lpage>-28</lpage>
</bibl>

<bibl id="B15">
  <title><p>Clinical natural language processing in languages other than
  english: opportunities and challenges</p></title>
  <aug>
    <au><snm>N{\'e}v{\'e}ol</snm><fnm>A</fnm></au>
    <au><snm>Dalianis</snm><fnm>H</fnm></au>
    <au><snm>Velupillai</snm><fnm>S</fnm></au>
    <au><snm>Savova</snm><fnm>G</fnm></au>
    <au><snm>Zweigenbaum</snm><fnm>P</fnm></au>
  </aug>
  <source>Journal of biomedical semantics</source>
  <publisher>Springer</publisher>
  <pubdate>2018</pubdate>
  <volume>9</volume>
  <issue>1</issue>
  <fpage>12</fpage>
</bibl>

<bibl id="B16">
  <title><p>Overview of the CLEF eHealth 2019 Multilingual Information
  Extraction</p></title>
  <aug>
    <au><snm>D{\"o}rendahl</snm><fnm>A</fnm></au>
    <au><snm>Leich</snm><fnm>N</fnm></au>
    <au><snm>Hummel</snm><fnm>B</fnm></au>
    <au><snm>Sch{\"o}nfelder</snm><fnm>G</fnm></au>
    <au><snm>Grune</snm><fnm>B</fnm></au>
  </aug>
  <pubdate>2019</pubdate>
</bibl>

<bibl id="B17">
  <title><p>GGPONC: A Corpus of German Medical Text with Rich Metadata Based on
  Clinical Practice Guidelines</p></title>
  <aug>
    <au><snm>Borchert</snm><fnm>F</fnm></au>
    <au><snm>Lohr</snm><fnm>C</fnm></au>
    <au><snm>Modersohn</snm><fnm>L</fnm></au>
    <au><snm>Langer</snm><fnm>T</fnm></au>
    <au><snm>Follmann</snm><fnm>M</fnm></au>
    <au><snm>Sachs</snm><fnm>JP</fnm></au>
    <au><snm>Hahn</snm><fnm>U</fnm></au>
    <au><snm>Schapranow</snm><fnm>MP</fnm></au>
  </aug>
  <source>Proceedings of the 11th International Workshop on Health Text Mining
  and Information Analysis</source>
  <pubdate>2020</pubdate>
  <fpage>38</fpage>
  <lpage>-48</lpage>
</bibl>

<bibl id="B18">
  <title><p>From Witch’s Shot to Music Making Bones-Resources for Medical
  Laymen to Technical Language and Vice Versa</p></title>
  <aug>
    <au><snm>Seiffe</snm><fnm>L</fnm></au>
    <au><snm>Marten</snm><fnm>O</fnm></au>
    <au><snm>Mikhailov</snm><fnm>M</fnm></au>
    <au><snm>Schmeier</snm><fnm>S</fnm></au>
    <au><snm>M{\"o}ller</snm><fnm>S</fnm></au>
    <au><snm>Roller</snm><fnm>R</fnm></au>
  </aug>
  <source>Proceedings of the 12th Language Resources and Evaluation
  Conference</source>
  <pubdate>2020</pubdate>
  <fpage>6185</fpage>
  <lpage>-6192</lpage>
</bibl>

<bibl id="B19">
  <title><p>Sharing Copies of Synthetic Clinical Corpora without Physical
  Distribution—A Case Study to Get Around IPRs and Privacy Constraints
  Featuring the German JSYNCC Corpus</p></title>
  <aug>
    <au><snm>Lohr</snm><fnm>C</fnm></au>
    <au><snm>Buechel</snm><fnm>S</fnm></au>
    <au><snm>Hahn</snm><fnm>U</fnm></au>
  </aug>
  <source>Proceedings of the Eleventh International Conference on Language
  Resources and Evaluation (LREC 2018)</source>
  <pubdate>2018</pubdate>
</bibl>

<bibl id="B20">
  <title><p>{Sharing Models and Tools for Processing German Clinical
  Texts}</p></title>
  <aug>
    <au><snm>Hellrich</snm><fnm>J</fnm></au>
    <au><snm>Matthies</snm><fnm>F</fnm></au>
    <au><snm>Faessler</snm><fnm>E</fnm></au>
    <au><snm>Hahn</snm><fnm>U</fnm></au>
  </aug>
  <source>MIE 2015 - Digital Healthcare Empowering Europeans</source>
  <publisher>Studies in Health Technology and Informatics</publisher>
  <pubdate>2015</pubdate>
  <volume>210</volume>
  <fpage>734</fpage>
  <lpage>-738</lpage>
</bibl>

<bibl id="B21">
  <title><p>Really, is medical sublanguage that different? Experimental
  counter-evidence from tagging medical and newspaper corpora.</p></title>
  <aug>
    <au><snm>Wermter</snm><fnm>J</fnm></au>
    <au><snm>Hahn</snm><fnm>U</fnm></au>
  </aug>
  <source>Medinfo</source>
  <pubdate>2004</pubdate>
  <fpage>560</fpage>
  <lpage>-564</lpage>
</bibl>

<bibl id="B22">
  <title><p>A Simple Algorithm for Identifying Negated Findings and Diseases in
  Discharge Summaries</p></title>
  <aug>
    <au><snm>Chapman</snm><fnm>WW</fnm></au>
    <au><snm>Bridewell</snm><fnm>W</fnm></au>
    <au><snm>Hanbury</snm><fnm>P</fnm></au>
    <au><snm>Cooper</snm><fnm>GF</fnm></au>
    <au><snm>Buchanan</snm><fnm>BG</fnm></au>
  </aug>
  <source>Journal of Biomedical Informatics</source>
  <pubdate>2001</pubdate>
  <volume>34</volume>
  <issue>5</issue>
  <fpage>301</fpage>
  <lpage>310</lpage>
</bibl>

<bibl id="B23">
  <title><p>Extending the NegEx lexicon for multiple languages</p></title>
  <aug>
    <au><snm>Chapman</snm><fnm>WW</fnm></au>
    <au><snm>Hilert</snm><fnm>D</fnm></au>
    <au><snm>Velupillai</snm><fnm>S</fnm></au>
    <au><snm>Kvist</snm><fnm>M</fnm></au>
    <au><snm>Skeppstedt</snm><fnm>M</fnm></au>
    <au><snm>Chapman</snm><fnm>BE</fnm></au>
    <au><snm>Conway</snm><fnm>M</fnm></au>
    <au><snm>Tharp</snm><fnm>M</fnm></au>
    <au><snm>Mowery</snm><fnm>DL</fnm></au>
    <au><snm>Deleger</snm><fnm>L</fnm></au>
  </aug>
  <source>Studies in health technology and informatics</source>
  <publisher>NIH Public Access</publisher>
  <pubdate>2013</pubdate>
  <volume>192</volume>
  <fpage>677</fpage>
</bibl>

<bibl id="B24">
  <title><p>Negation Detection in Clinical Reports Written in
  {G}erman</p></title>
  <aug>
    <au><snm>Cotik</snm><fnm>V</fnm></au>
    <au><snm>Roller</snm><fnm>R</fnm></au>
    <au><snm>Xu</snm><fnm>F</fnm></au>
    <au><snm>Uszkoreit</snm><fnm>H</fnm></au>
    <au><snm>Budde</snm><fnm>K</fnm></au>
    <au><snm>Schmidt</snm><fnm>D</fnm></au>
  </aug>
  <source>Proceedings of the Fifth Workshop on Building and Evaluating
  Resources for Biomedical Text Mining ({B}io{T}xt{M}2016)</source>
  <publisher>Osaka, Japan: The COLING 2016 Organizing Committee</publisher>
  <pubdate>2016</pubdate>
  <fpage>115</fpage>
  <lpage>-124</lpage>
</bibl>

<bibl id="B25">
  <title><p>{A Domain-adapted Dependency Parser for German Clinical
  Text}</p></title>
  <aug>
    <au><snm>Kara</snm><fnm>E</fnm></au>
    <au><snm>Zeen</snm><fnm>T</fnm></au>
    <au><snm>Gabryszak</snm><fnm>A</fnm></au>
    <au><snm>Budde</snm><fnm>K</fnm></au>
    <au><snm>Schmidt</snm><fnm>D</fnm></au>
    <au><snm>Roller</snm><fnm>R</fnm></au>
  </aug>
  <source>Proceedings of the 14th Conference on Natural Language Processing
  (KONVENS 2018)</source>
  <publisher>Vienna, Austria</publisher>
  <pubdate>2018</pubdate>
</bibl>

<bibl id="B26">
  <title><p>Unsupervised Abbreviation Expansion in Clinical
  Narratives.</p></title>
  <aug>
    <au><snm>Oleynik</snm><fnm>M</fnm></au>
    <au><snm>Kreuzthaler</snm><fnm>M</fnm></au>
    <au><snm>Schulz</snm><fnm>S</fnm></au>
  </aug>
  <source>MedInfo</source>
  <pubdate>2017</pubdate>
  <volume>245</volume>
  <fpage>539</fpage>
  <lpage>-543</lpage>
</bibl>

<bibl id="B27">
  <title><p>{Pseudonymization of PHI items in German clinical
  reports}</p></title>
  <aug>
    <au><snm>Lohr</snm><fnm>C</fnm></au>
    <au><snm>Eder</snm><fnm>E</fnm></au>
    <au><snm>Hahn</snm><fnm>U</fnm></au>
  </aug>
  <source>Public Health and Informatics</source>
  <pubdate>2021</pubdate>
</bibl>

<bibl id="B28">
  <title><p>Annotation and initial evaluation of a large annotated German
  oncological corpus</p></title>
  <aug>
    <au><snm>Kittner</snm><fnm>M</fnm></au>
    <au><snm>Lamping</snm><fnm>M</fnm></au>
    <au><snm>Rieke</snm><fnm>DT</fnm></au>
    <au><snm>G{\"o}tze</snm><fnm>J</fnm></au>
    <au><snm>Bajwa</snm><fnm>B</fnm></au>
    <au><snm>Jelas</snm><fnm>I</fnm></au>
    <au><snm>R{\"u}ter</snm><fnm>G</fnm></au>
    <au><snm>Hautow</snm><fnm>H</fnm></au>
    <au><snm>S{\"a}nger</snm><fnm>M</fnm></au>
    <au><snm>Habibi</snm><fnm>M</fnm></au>
    <au><cnm>others</cnm></au>
  </aug>
  <source>JAMIA open</source>
  <publisher>Oxford University Press</publisher>
  <pubdate>2021</pubdate>
  <volume>4</volume>
  <issue>2</issue>
  <fpage>ooab025</fpage>
</bibl>

<bibl id="B29">
  <title><p>GERNERMED: An open German medical NER model</p></title>
  <aug>
    <au><snm>Frei</snm><fnm>J</fnm></au>
    <au><snm>Kramer</snm><fnm>F</fnm></au>
  </aug>
  <source>Software Impacts</source>
  <publisher>Elsevier</publisher>
  <pubdate>2022</pubdate>
  <volume>11</volume>
  <fpage>100212</fpage>
</bibl>

<bibl id="B30">
  <title><p>2018 n2c2 shared task on adverse drug events and medication
  extraction in electronic health records</p></title>
  <aug>
    <au><snm>Henry</snm><fnm>S</fnm></au>
    <au><snm>Buchan</snm><fnm>K</fnm></au>
    <au><snm>Filannino</snm><fnm>M</fnm></au>
    <au><snm>Stubbs</snm><fnm>A</fnm></au>
    <au><snm>Uzuner</snm><fnm>O</fnm></au>
  </aug>
  <source>Journal of the American Medical Informatics Association</source>
  <publisher>Oxford University Press</publisher>
  <pubdate>2020</pubdate>
  <volume>27</volume>
  <issue>1</issue>
  <fpage>3</fpage>
  <lpage>-12</lpage>
</bibl>

<bibl id="B31">
  <title><p>Development of a language model for medical domain</p></title>
  <aug>
    <au><snm>Shrestha</snm><fnm>M</fnm></au>
  </aug>
  <source>Master's thesis</source>
  <publisher>Hochschule Rhein-Waal</publisher>
  <pubdate>2021</pubdate>
</bibl>

<bibl id="B32">
  <title><p>German-MedBERT on Hugging Face</p></title>
  <source>\url{https://huggingface.co/smanjil/German-MedBERT}</source>
  <note>Accessed: 2022-03-15</note>
</bibl>

<bibl id="B33">
  <title><p>Semi-automated De-identification of German Content Sensitive
  Reports for Big Data Analytics</p></title>
  <aug>
    <au><snm>Seuss</snm><fnm>H</fnm></au>
    <au><snm>Dankerl</snm><fnm>P</fnm></au>
    <au><snm>Ihle</snm><fnm>M</fnm></au>
    <au><snm>Grandjean</snm><fnm>A</fnm></au>
    <au><snm>Hammon</snm><fnm>R</fnm></au>
    <au><snm>Kaestle</snm><fnm>N</fnm></au>
    <au><snm>Fasching</snm><fnm>PA</fnm></au>
    <au><snm>Maier</snm><fnm>C</fnm></au>
    <au><snm>Christoph</snm><fnm>J</fnm></au>
    <au><snm>Sedlmayr</snm><fnm>M</fnm></au>
    <au><cnm>others</cnm></au>
  </aug>
  <source>R{\"o}Fo-Fortschritte auf dem Gebiet der R{\"o}ntgenstrahlen und der
  bildgebenden Verfahren</source>
  <pubdate>2017</pubdate>
  <volume>189</volume>
  <issue>07</issue>
  <fpage>661</fpage>
  <lpage>-671</lpage>
</bibl>

<bibl id="B34">
  <title><p>Beckers Abkürzungslexikon Medizinischer Begriffe</p></title>
  <source>\url{https://www.medizinische-abkuerzungen.de/suche.html}</source>
  <note>Accessed: 2021-01-10</note>
</bibl>

<bibl id="B35">
  <title><p>A fine-grained corpus annotation schema of German nephrology
  records</p></title>
  <aug>
    <au><snm>Roller</snm><fnm>R</fnm></au>
    <au><snm>Uszkoreit</snm><fnm>H</fnm></au>
    <au><snm>Xu</snm><fnm>F</fnm></au>
    <au><snm>Seiffe</snm><fnm>L</fnm></au>
    <au><snm>Mikhailov</snm><fnm>M</fnm></au>
    <au><snm>Staeck</snm><fnm>O</fnm></au>
    <au><snm>Budde</snm><fnm>K</fnm></au>
    <au><snm>Halleck</snm><fnm>F</fnm></au>
    <au><snm>Schmidt</snm><fnm>D</fnm></au>
  </aug>
  <source>Proceedings of the Clinical Natural Language Processing Workshop
  (ClinicalNLP)</source>
  <publisher>Osaka, Japan: The COLING 2016 Organizing Committee</publisher>
  <pubdate>2016</pubdate>
  <fpage>69</fpage>
  <lpage>-77</lpage>
</bibl>

<bibl id="B36">
  <title><p>{brat}: a Web-based Tool for {NLP}-Assisted Text
  Annotation</p></title>
  <aug>
    <au><snm>Stenetorp</snm><fnm>P</fnm></au>
    <au><snm>Pyysalo</snm><fnm>S</fnm></au>
    <au><snm>Topi\'{c}</snm><fnm>G</fnm></au>
    <au><snm>Ohta</snm><fnm>T</fnm></au>
    <au><snm>Ananiadou</snm><fnm>S</fnm></au>
    <au><snm>Tsujii</snm><fnm>J</fnm></au>
  </aug>
  <source>Proceedings of the Demonstrations Session at {EACL} 2012</source>
  <publisher>Avignon, France: Association for Computational
  Linguistics</publisher>
  <pubdate>2012</pubdate>
</bibl>

<bibl id="B37">
  <title><p>Agreement, the f-measure, and reliability in information
  retrieval</p></title>
  <aug>
    <au><snm>Hripcsak</snm><fnm>G</fnm></au>
    <au><snm>Rothschild</snm><fnm>AS</fnm></au>
  </aug>
  <source>Journal of the American medical informatics association</source>
  <publisher>BMJ Group BMA House, Tavistock Square, London, WC1H
  9JR</publisher>
  <pubdate>2005</pubdate>
  <volume>12</volume>
  <issue>3</issue>
  <fpage>296</fpage>
  <lpage>-298</lpage>
</bibl>

<bibl id="B38">
  <title><p>{Linguistic Modeling for Text Analytic Tasks for German Clinical
  Texts}</p></title>
  <aug>
    <au><snm>Seiffe</snm><fnm>L</fnm></au>
  </aug>
  <source>Master's thesis</source>
  <publisher>TU Berlin</publisher>
  <pubdate>2018</pubdate>
</bibl>

<bibl id="B39">
  <title><p>{Guidelines f{\"u}r das Tagging deutscher Textcorpora mit
  STTS}</p></title>
  <aug>
    <au><snm>Schiller</snm><fnm>A</fnm></au>
    <au><snm>Teufel</snm><fnm>S</fnm></au>
    <au><snm>Thielen</snm><fnm>C</fnm></au>
  </aug>
  <source>Universit{\"a}ten Stuttgart und T{\"u}bingen</source>
  <pubdate>1999</pubdate>
</bibl>

<bibl id="B40">
  <title><p>Universal dependencies v1: A multilingual treebank
  collection</p></title>
  <aug>
    <au><snm>Nivre</snm><fnm>J</fnm></au>
    <au><snm>De Marneffe</snm><fnm>MC</fnm></au>
    <au><snm>Ginter</snm><fnm>F</fnm></au>
    <au><snm>Goldberg</snm><fnm>Y</fnm></au>
    <au><snm>Hajic</snm><fnm>J</fnm></au>
    <au><snm>Manning</snm><fnm>CD</fnm></au>
    <au><snm>McDonald</snm><fnm>R</fnm></au>
    <au><snm>Petrov</snm><fnm>S</fnm></au>
    <au><snm>Pyysalo</snm><fnm>S</fnm></au>
    <au><snm>Silveira</snm><fnm>N</fnm></au>
    <au><cnm>others</cnm></au>
  </aug>
  <source>Proceedings of the Tenth International Conference on Language
  Resources and Evaluation (LREC'16)</source>
  <pubdate>2016</pubdate>
  <fpage>1659</fpage>
  <lpage>-1666</lpage>
</bibl>

<bibl id="B41">
  <title><p>Because Size Does Matter: The Hamburg Dependency
  Treebank</p></title>
  <aug>
    <au><snm>Foth</snm><fnm>KA</fnm></au>
    <au><snm>Köhn</snm><fnm>A</fnm></au>
    <au><snm>Beuck</snm><fnm>N</fnm></au>
    <au><snm>Menzel</snm><fnm>W</fnm></au>
  </aug>
  <source>Proceedings of the Ninth International Conference on Language
  Resources and Evaluation (LREC'14)</source>
  <publisher>Reykjavik, Iceland: European Language Resources Association
  (ELRA)</publisher>
  <editor>Nicoletta Calzolari (Conference Chair) and Khalid Choukri and Thierry
  Declerck and Hrafn Loftsson and Bente Maegaard and Joseph Mariani and
  Asuncion Moreno and Jan Odijk and Stelios Piperidis</editor>
  <pubdate>2014</pubdate>
</bibl>

<bibl id="B42">
  <title><p>Med1 Forum</p></title>
  <source>\url{https://med2-forum.de/}</source>
  <note>Accessed: 2021-01-10; Note, the Med1 forum does not exist anymore, see
  med2 instead.</note>
</bibl>

<bibl id="B43">
  <title><p>Deutsches Medizin Forum</p></title>
  <source>\url{https://www.medizin-forum.de/}</source>
  <note>Accessed: 2021-01-10</note>
</bibl>

<bibl id="B44">
  <title><p>Spiegel</p></title>
  <source>\url{https://www.spiegel.de/gesundheit/}</source>
  <note>Accessed: 2021-01-10</note>
</bibl>

<bibl id="B45">
  <title><p>Aerzte-Blatt</p></title>
  <source>\url{https://www.aerzteblatt.de/}</source>
  <note>Accessed: 2021-01-10</note>
</bibl>

<bibl id="B46">
  <title><p>NetDoktor</p></title>
  <source>\url{https://www.netdoktor.de/}</source>
  <note>Accessed: 2021-01-10</note>
</bibl>

<bibl id="B47">
  <title><p>Onmeda</p></title>
  <source>\url{https://www.onmeda.de/}</source>
  <note>Accessed: 2021-01-10</note>
</bibl>

<bibl id="B48">
  <title><p>German PubMed Abstracts</p></title>
  <source>\url{https://pubmed.ncbi.nlm.nih.gov/}</source>
  <note>Accessed: 2021-01-10</note>
</bibl>

<bibl id="B49">
  <title><p>eDocTrainer</p></title>
  <source>\url{https://www.edoctrainer.de/}</source>
  <note>Accessed: 2021-01-10</note>
</bibl>

<bibl id="B50">
  <title><p>Chirurgie: mit integriertem Fallquiz - 40 F{\"a}lle nach neuer
  AO</p></title>
  <aug>
    <au><snm>Siewert</snm><fnm>J.R.</fnm></au>
    <au><snm>Stein</snm><fnm>H.J.</fnm></au>
    <au><snm>Allg{\"o}wer</snm><fnm>M.</fnm></au>
  </aug>
  <publisher>Springer Verlag Berlin Heidelberg</publisher>
  <pubdate>2009</pubdate>
</bibl>

<bibl id="B51">
  <title><p>Neurologie</p></title>
  <aug>
    <au><snm>Poeck</snm><fnm>K.</fnm></au>
    <au><snm>Hacke</snm><fnm>W.</fnm></au>
  </aug>
  <publisher>Springer Verlag Berlin Heidelberg</publisher>
  <pubdate>2006</pubdate>
</bibl>

<bibl id="B52">
  <title><p>Urologie</p></title>
  <aug>
    <au><snm>Hautmann</snm><fnm>R.</fnm></au>
    <au><snm>Gschwend</snm><fnm>J.E.</fnm></au>
  </aug>
  <publisher>Springer Verlag Berlin Heidelberg</publisher>
  <pubdate>2014</pubdate>
</bibl>

<bibl id="B53">
  <title><p>Basiswissen Augenheilkunde</p></title>
  <aug>
    <au><snm>Walter</snm><fnm>P</fnm></au>
    <au><snm>Plange</snm><fnm>N</fnm></au>
  </aug>
  <publisher>Springer</publisher>
  <pubdate>2017</pubdate>
</bibl>

<bibl id="B54">
  <title><p>Hals-Nasen-Ohren-Heilkunde</p></title>
  <aug>
    <au><snm>Lenarz</snm><fnm>T.</fnm></au>
    <au><snm>Boenninghaus</snm><fnm>H.G.</fnm></au>
  </aug>
  <publisher>Springer Verlag Berlin Heidelberg</publisher>
  <pubdate>2012</pubdate>
</bibl>

<bibl id="B55">
  <title><p>Notfallmedizin</p></title>
  <aug>
    <au><snm>Ziegenfu{\ss}</snm><fnm>T.</fnm></au>
  </aug>
  <publisher>Springer Verlag Berlin Heidelberg</publisher>
  <pubdate>2016</pubdate>
</bibl>

<bibl id="B56">
  <title><p>Basiswissen Dermatologie</p></title>
  <aug>
    <au><snm>Goebeler</snm><fnm>M.</fnm></au>
    <au><snm>Hamm</snm><fnm>H.</fnm></au>
  </aug>
  <publisher>Springer Verlag Berlin Heidelberg</publisher>
  <pubdate>2017</pubdate>
</bibl>

<bibl id="B57">
  <title><p>Basiswissen Psychiatrie und Psychotherapie</p></title>
  <aug>
    <au><snm>Arolt</snm><fnm>V</fnm></au>
    <au><snm>Reimer</snm><fnm>C</fnm></au>
    <au><snm>Dilling</snm><fnm>H</fnm></au>
  </aug>
  <publisher>Springer-Verlag</publisher>
  <pubdate>2011</pubdate>
</bibl>

<bibl id="B58">
  <title><p>Enriching word vectors with subword information</p></title>
  <aug>
    <au><snm>Bojanowski</snm><fnm>P</fnm></au>
    <au><snm>Grave</snm><fnm>E</fnm></au>
    <au><snm>Joulin</snm><fnm>A</fnm></au>
    <au><snm>Mikolov</snm><fnm>T</fnm></au>
  </aug>
  <source>Transactions of the association for computational
  linguistics</source>
  <publisher>MIT Press</publisher>
  <pubdate>2017</pubdate>
  <volume>5</volume>
  <fpage>135</fpage>
  <lpage>-146</lpage>
</bibl>

<bibl id="B59">
  <title><p>Contextual String Embeddings for Sequence Labeling</p></title>
  <aug>
    <au><snm>Akbik</snm><fnm>A</fnm></au>
    <au><snm>Blythe</snm><fnm>D</fnm></au>
    <au><snm>Vollgraf</snm><fnm>R</fnm></au>
  </aug>
  <source>Proceedings of the 27th International Conference on Computational
  Linguistics</source>
  <publisher>Santa Fe, New Mexico, USA: Association for Computational
  Linguistics</publisher>
  <pubdate>2018</pubdate>
  <fpage>1638</fpage>
  <lpage>-1649</lpage>
</bibl>

<bibl id="B60">
  <title><p>Pooled Contextualized Embeddings for Named Entity
  Recognition</p></title>
  <aug>
    <au><snm>Akbik</snm><fnm>A</fnm></au>
    <au><snm>Bergmann</snm><fnm>T</fnm></au>
    <au><snm>Vollgraf</snm><fnm>R</fnm></au>
  </aug>
  <source>Proceedings of the 2019 Conference of the North {A}merican Chapter of
  the Association for Computational Linguistics: Human Language Technologies,
  Volume 1 (Long and Short Papers)</source>
  <publisher>Minneapolis, Minnesota: Association for Computational
  Linguistics</publisher>
  <pubdate>2019</pubdate>
  <fpage>724</fpage>
  <lpage>-728</lpage>
</bibl>

<bibl id="B61">
  <title><p>Relation extraction: Perspective from convolutional neural
  networks</p></title>
  <aug>
    <au><snm>Nguyen</snm><fnm>TH</fnm></au>
    <au><snm>Grishman</snm><fnm>R</fnm></au>
  </aug>
  <source>Proceedings of the 1st Workshop on Vector Space Modeling for Natural
  Language Processing</source>
  <pubdate>2015</pubdate>
  <fpage>39</fpage>
  <lpage>-48</lpage>
</bibl>

<bibl id="B62">
  <title><p>What's in a Drug Name?: A rose might smell as sweet by any other
  name, but the process of naming the growing number of medications has become
  quite complex and serious.</p></title>
  <aug>
    <au><snm>Wick</snm><fnm>JY</fnm></au>
  </aug>
  <source>Journal of the American Pharmacists Association</source>
  <pubdate>2004</pubdate>
  <volume>44</volume>
  <issue>1</issue>
  <fpage>12</fpage>
  <lpage>-14</lpage>
</bibl>

</refgrp>
} 
\end{backmatter}
\end{document}